\documentclass{article} % For LaTeX2e
\PassOptionsToPackage{table}{xcolor}
\usepackage{iclr2027_conference,times}

\usepackage{amsmath,amsfonts,bm}

\def\eqref#1{equation~\ref{#1}}
\def\1{\bm{1}}

\DeclareMathAlphabet{\mathsfit}{\encodingdefault}{\sfdefault}{m}{sl}
\SetMathAlphabet{\mathsfit}{bold}{\encodingdefault}{\sfdefault}{bx}{n}

\usepackage[utf8]{inputenc}
\usepackage[T1]{fontenc}
\makeatletter
\let\@check@eq\@gobbletwo
\makeatother
\usepackage{amsmath,amssymb,amsfonts,amsthm,mathtools,bm}

\usepackage{booktabs,multirow,array,tabularx}
\usepackage{graphicx}
\usepackage{subcaption}
\usepackage{xcolor}
\usepackage{enumitem}
\usepackage{hyperref}
\usepackage[nameinlink,capitalize]{cleveref}
\usepackage{algorithm}
\usepackage{algorithmic}
\usepackage{float}
\usepackage{url}
\usepackage{pifont}
\usepackage{wrapfig}
\usepackage{needspace}
\usepackage[most]{tcolorbox}
\newcommand{\meanstd}[2]{\ensuremath{#1{\scriptstyle\,\pm #2}}}
\newcommand{\bestmeanstd}[2]{\ensuremath{\mathbf{#1}{\scriptstyle\,\pm #2}}}
\newcommand{\secondmeanstd}[2]{\ensuremath{\underline{#1}{\scriptstyle\,\pm #2}}}

\usepackage{wrapfig}
\usepackage{tabularx}
\usepackage{array}
\usepackage[table]{xcolor}
\hypersetup{
  colorlinks=true,
  linkcolor=blue!60!black,
  citecolor=blue!60!black,
  urlcolor=blue!60!black,
  pdftitle={Act with Intent: Distilling Behavior Intent for Vision-Language-Action Models},
  pdfauthor={Sangoh Lee, Sangwoo Mo, Wook-Shin Han}
}

\DeclareRobustCommand{\method}{\textnormal{\textsc{Indi}}}
\DeclareRobustCommand{\methodfull}{Intention Distillation}

\newcommand{\squishlist}{
   \begin{list}{$\bullet$}
    { \setlength{\itemsep}{1pt}
      \setlength{\parsep}{0pt}
      \setlength{\topsep}{2pt}
      \setlength{\partopsep}{0pt}
      \setlength{\listparindent}{-2pt}
      \setlength{\itemindent}{-5pt}
      \setlength{\leftmargin}{1.5em}
      \setlength{\labelwidth}{0em}
      \setlength{\labelsep}{0.5em}
    }
}
\newcommand{\squishlistend}{
    \end{list}  }

\title{Act with Intent: Distilling Behavior Intent for Vision-Language-Action Models}

\iclrpreprintcopy
\author{
Sangoh Lee$^{1}$, \quad
Sangwoo Mo$^{2,\ast}$, \quad
Wook-Shin Han$^{1,\ast}$ \\
$^{1}$GSAI, POSTECH \quad
$^{2}$IME, POSTECH \\
{\small
$^{1}$\texttt{\{solee,wshan\}@dblab.postech.ac.kr}
\quad
$^{2}$\texttt{sangwoo.mo@postech.ac.kr}
}
}

\begin{document}
\raggedbottom

\begingroup
\renewcommand{\thefootnote}{\fnsymbol{footnote}}
\footnotetext[1]{Corresponding authors.}
\endgroup

\maketitle

\begin{abstract}
Vision-Language-Action (VLA) models can turn multimodal context into robot actions, but their action decoders are still trained largely by behavior cloning.
This supervises which motor command was demonstrated while leaving implicit the local objective served by the behavior under the instruction.
Future-based supervision enriches action learning with frames, latent observations, trajectories, or motion representations, but these signals capture particular realizations of what may happen rather than the shared semantic objective of the forthcoming behavior.
We propose \methodfull{} (\method{}), which distills behavior-level intent into the action decoder.
During training, a frozen teacher VLM interprets a demonstrated segment from the current observation, instruction, coarse action summary, and corresponding execution video.
From its standard inputs, the deployed VLA recovers the resulting multimodal intent representation at an intermediate decoder layer and uses it to organize action prediction together with representations of how the behavior unfolds and what it achieves.
On SimplerEnv-Bridge, \method{} improves GR00T-N1.7 from \(64.3\%\) to \(84.7\%\), and on RoboCasa Kitchen it improves the controlled GR00T-N1.7 baseline from \(64.1\%\) to \(70.3\%\), with consistent gains on \(\pi_{0.5}\) across both benchmarks.
In real-world tasks, \method{} improves average success from \(62.0\%\) to \(68.7\%\), with gains of up to \(12.0\) pp on longer-horizon tasks.
Further analyses show that the recovered latent is used by the decoder, captures behavior objective and execution progress, and organizes downstream predictions in an objective-dependent manner.
These results show that action decoders benefit from explicitly modeling the semantic objective of the behavior they generate.
Project page: \url{https://leesangoh.github.io/indi-project-page/}
\end{abstract}

\section{Introduction}
\label{sec:intro}
% Recent Vision-Language-Action (VLA) models and robot foundation models have advanced rapidly by scaling vision-language backbones, action experts, and robot demonstrations \citep{zitkovich2023rt2,kim2025openvla,black2025pi0,black2025pi05,physicalintelligence2026pi07,bjorck2025grootn1,nvidia2026grootn17,geminirobotics2025,li2024cogact}. These models show that Internet-scale visual and linguistic knowledge can be transferred to robot control, and that specialized action heads can turn multimodal context into continuous motor commands. Yet the dominant supervision at the action level remains narrow. 
% Behavior cloning specifies which action was taken under a given observation and instruction, but it does not explicitly teach what the action is meant to accomplish. 
% As a result, an action decoder may learn a strong context-to-motion mapping without forming a representation that organizes the action around its functional purpose.

Recent Vision-Language-Action (VLA) models and robot foundation models have advanced rapidly by scaling vision-language backbones, action decoders, and robot demonstrations \citep{zitkovich2023rt2,kim2025openvla,black2025pi0,black2025pi05,physicalintelligence2026pi07,bjorck2025grootn1,nvidia2026grootn17,geminirobotics2025,li2024cogact}. These models show that Internet-scale visual and linguistic knowledge can be transferred to robot control, and that specialized action decoders can turn multimodal context into continuous motor commands. Yet action-level training remains largely imitation-based. 
Behavior cloning tells the decoder which motor command to reproduce for a given observation and instruction, but not what that behavior is supposed to achieve under the instruction.
As a result, the decoder can learn a strong context-to-action mapping while leaving the purpose of each behavior implicit.

To go beyond action-only supervision, a growing line of work augments VLA policies with future-based supervision.
Some methods use future scene states, including generated frames, subgoal images, and latent observations \citep{wu2024gr1,black2023susie,hu2025vpp,zheng2025flare}.
Others represent how the forthcoming interaction unfolds through trajectories, motion fields, point tracks, or structured world dynamics \citep{zhao2025cotvla,zhang2025dreamvla,xu2024im2flow2act,bharadhwaj2024track2act}.
These signals provide valuable information about what the future may look like or how motion may unfold, but they supervise particular realizations of behavior rather than the objective those realizations serve.
This distinction matters because similar stages of manipulation can serve different objectives across tasks, while the same objective can be realized through diverse executions.
To give the forthcoming action sequence its meaning under the instruction, the action decoder must represent the local objective it serves.
This objective guides the decoder in translating the scene and instruction context encoded by the VLA's vision-language module into coherent, temporally extended behavior.
We therefore argue that the decoder should recover a behavior-level intent that organizes action together with diverse representations of how the behavior unfolds and what it achieves.

\begin{figure}[htbp]
\centering
\includegraphics[width=\linewidth]{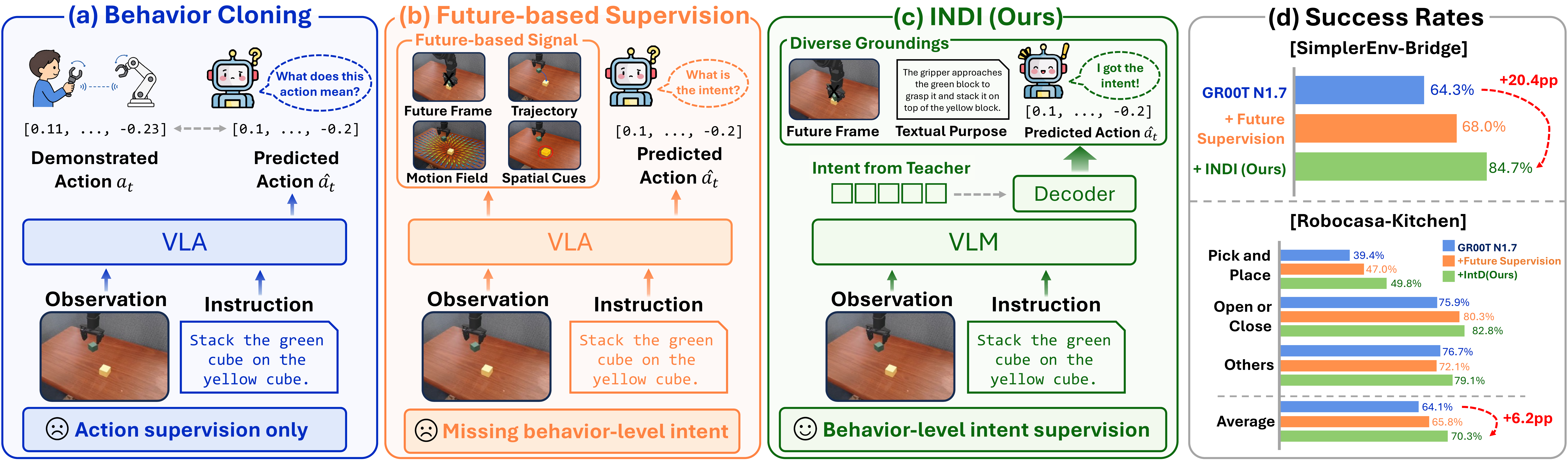}
\caption{\textbf{From behavior cloning to intent distillation.}
(a) Behavior cloning directly supervises actions.
(b) Future-based supervision adds representations of future states or motion.
(c) \method{} distills behavior-level intent from executed behavior into the action decoder, where it organizes action and representations of how the behavior unfolds and what it achieves.
Right: GR00T-N1.7 results on SimplerEnv-Bridge and RoboCasa.}
\label{fig:teaser}
\vspace{-8pt}
\end{figure}

To operationalize this principle, we propose \methodfull{} (\method{}), which distills behavior-level intent into the action decoder.
We define intent as the local objective that the forthcoming behavior should achieve under the instruction.
During training, a frozen teacher VLM interprets an executed behavior segment from the current observation, instruction, a coarse action summary, and the execution video.
The student decoder learns to recover the resulting multimodal intent representation from the current observation, instruction, and proprioceptive state, and uses it as an intermediate semantic state for action prediction.

Concretely, \method{} realizes this behavior-level intent supervision through three components.
\textbf{(1) Intent from executed behavior.}
The teacher VLM identifies what object or relation changes, what local objective the segment serves, and how it advances the instruction.
We use the teacher's multimodal representation formed during this interpretation as the intent target, together with a generated purpose statement and an endpoint visual feature as complementary groundings.
\textbf{(2) Intent recovery inside the action decoder.}
Learnable intent queries recover the teacher target at an intermediate decoder layer, after which the remaining layers jointly complete action, latent visual and textual grounding predictions.
\textbf{(3) Intent-aware decoding.}
The recovered intent participates in action and grounding prediction, making it a functional intermediate representation rather than an auxiliary alignment target.
At deployment, the teacher and all target-generation modules are removed.

We evaluate whether behavior-level intent improves action decoding and whether the recovered representation exhibits the expected structure of intent.
On SimplerEnv-Bridge~\citep{li2025simpler}, \method{} improves GR00T-N1.7 from 64.3\% to 84.7\%.
On RoboCasa Kitchen~\citep{nasiriany2024robocasa}, it improves the controlled GR00T-N1.7 baseline from 64.1\% to 70.3\% across 24 tasks, with consistent gains on \(\pi_{0.5}\) across both benchmarks.
In real-world tasks, \method{} improves success from 62.0\% to 68.7\% and generalizes to held-out objects and distractors.
Representation analyses and closed-loop interventions further show that the recovered intent is used by the decoder, captures behavior objective and execution progress, and organizes downstream predictions in an objective-dependent manner.
Together, these results show that behavior-level semantic supervision improves policy performance and generalization while inducing an internal representation that functions as intent.

\noindent\textbf{Contributions.}
Our contributions are as follows:
%\squishlist
\begin{itemize}
    \item We identify behavior-level intent as a missing supervision target for VLA action decoders.
    Beyond imitating actions or predicting future states and motion, the decoder should model the local objective served by its action sequence.

    \item We propose \method{}, which distills a teacher VLM's multimodal understanding of executed behavior into intermediate action-decoder states.
    The recovered intent organizes action prediction and representations of how the behavior unfolds and what it achieves, with no teacher-side modules at deployment.

    \item Across VLA backbones, SimplerEnv-Bridge~\citep{li2025simpler}, RoboCasa~\citep{nasiriany2024robocasa}, and real-world manipulation, \method{} improves performance and OOD generalization.
    Analyses and interventions further show that the recovered latent functions as intent by encoding behavior objective and progress and organizing downstream predictions.
%\squishlistend
\end{itemize}

\vspace{-0.5cm}
\section{Related Work}

\noindent\textbf{VLA models.}
VLA models combine large multimodal backbones with robot demonstrations to map observations and instructions to actions \citep{brohan2023rt1,zitkovich2023rt2,kim2025openvla,octo2024,black2025pi0,black2025pi05,physicalintelligence2026pi07,bjorck2025grootn1,nvidia2026grootn17,geminirobotics2025,li2024cogact}.
Their progress has been supported by large robot datasets spanning diverse tasks, scenes, and embodiments \citep{openx2023,khazatsky2024droid,walke2023bridgedata,fang2023rh20t}.
Recent systems increasingly pair pretrained vision-language modules with action decoders trained using flow-matching or diffusion objectives \citep{black2025pi0,black2025pi05,bjorck2025grootn1,li2024cogact,liu2024rdt,wen2025diffusionvla,kim2026rldx}.
While these advances improve action modeling, decoder supervision remains dominated by imitation of demonstrated motor commands.
\method{} instead supervises the local objective served by the action sequence.

\noindent\textbf{Future-based and structured supervision.}
Prior work augments robot policies with generated future frames, subgoal images, and visual reasoning frames \citep{wu2024gr1,black2023susie,zhao2025cotvla}.
Other methods learn predictive visual features, future latent states, world-action representations, or structured world knowledge as intermediate signals for action decoding \citep{hu2025vpp,tian2025seer,zheng2025flare,zhang2025dreamvla,xu2026futurevla,sun2026vlajepa,beingbeyond2026beingh07,won2026dust}.
Related approaches represent spatial, motion, or semantic structure through optical flow, point tracks, visual traces, grounding masks, affordance chains, and embodied reasoning traces \citep{xu2024im2flow2act,bharadhwaj2024track2act,zheng2025tracevla,huang2025roboground,zawalski2025ecot,huang2025thinkact,li2025coavla}.
These methods demonstrate the value of intermediate supervision beyond motor commands, but define the decoder interface through a particular future state, motion structure, or reasoning representation.
\method{} instead supervises the behavior-level objective that organizes action and its diverse realizations.

\noindent\textbf{Intent and latent plans in robot policies.}
Related methods derive deployment-time latent plans from trajectories, with Play-LMP learning continuous plans from play and LADS learning language-regularized discrete plans from trajectory segments \citep{lynch2020playlmp,jiang2025lads}.
Their latents are reconstruction codes learned jointly with the policy, whereas \method{} distills a fixed teacher's semantic interpretation of the behavior.
Prior work also operationalizes intent through trajectory abstractions, action priors, and virtual targets \citep{huang2026mint,zhong2026acotvla,pang2026index}, through temporal context or predicted scene structure \citep{lian2026intentvla,chen2026dial,fan2026aim,xu2026futurevla}, or through human cues such as demonstrations, gaze, and indirect instructions \citep{gupta2026lucid,xie2026humanintentionpriors,tay2026intentglance,li2026gazevla,pani2026gazeregularizedvla,zuo2026gaze2act,chen2025intentionvla}.
\method{} instead defines intent as the local objective served by executed behavior under the instruction.
A training-only teacher derives a multimodal intent target from the segment, distilled into intermediate decoder states and grounded in visual outcome and textual purpose.
The deployed policy recovers it from standard VLA inputs.
\begin{figure}[t]
\centering
\includegraphics[width=\linewidth]{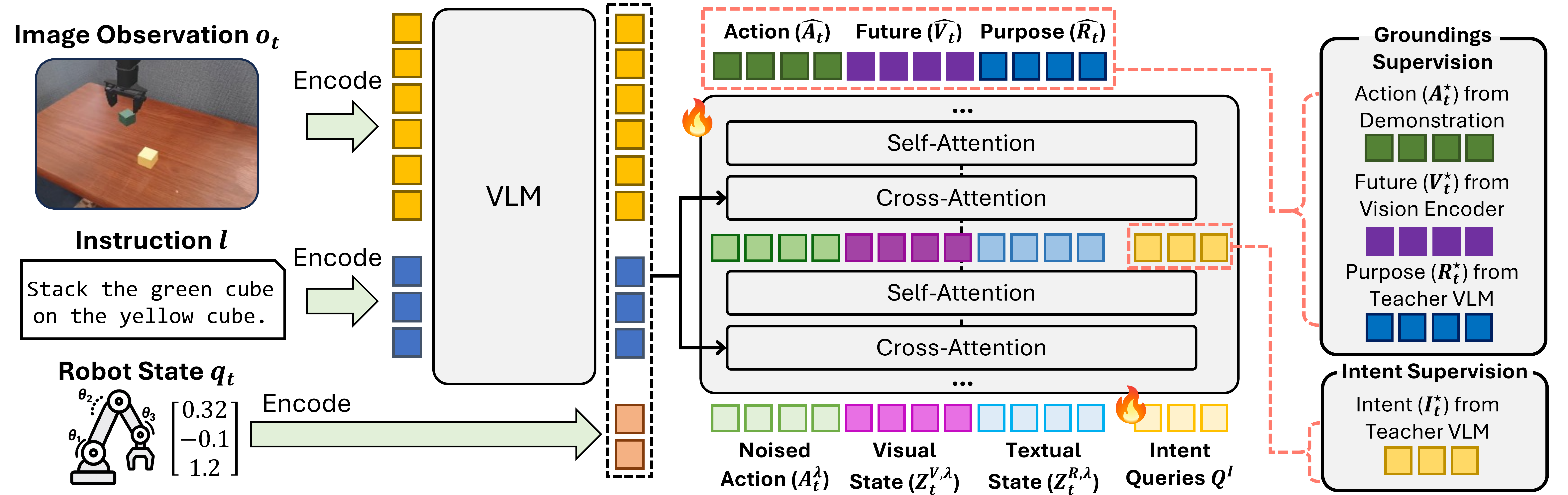}
\caption{\textbf{Overview of \method{}.}
A training-only teacher VLM derives multimodal intent and textual-purpose targets from executed behavior, while a frozen visual encoder supplies the endpoint-visual target.
The VLA decoder recovers intent at an intermediate layer and jointly predicts actions and latent visual and textual groundings, denoted by \(\widehat{V}_t\) and \(\widehat{R}_t\).
Dashed lines indicate training-only supervision, and all target-generation modules are removed at deployment.}
\label{fig:architecture}
\vspace{-0.5cm}
\end{figure}

\vspace{-8pt}
\section{Methodology}
\label{sec:method}
\vspace{-5pt}

We propose \method{}, which distills behavior-level intent into a pretrained VLA action decoder.
During training, a frozen teacher VLM interprets multimodal evidence of the demonstrated behavior under the instruction, producing an intent target.
The decoder recovers this target at an intermediate layer from its standard inputs and uses the recovered intent in subsequent action and visual and textual grounding prediction.
The overall framework is illustrated in Figure~\ref{fig:architecture}.

\vspace{-0.1cm}

\subsection{Problem Formulation}

We consider a pretrained VLA policy with a frozen vision-language module
\(\mathcal{F}_{\mathrm{vlm}}\) and a trainable flow-matching action decoder.
At time \(t\), the VLA receives an RGB observation \(o_t\), a language instruction \(\ell\), and a proprioceptive state \(q_t\).
The VLM produces context tokens \(B_t=\mathcal{F}_{\mathrm{vlm}}(o_t,\ell)\).

Our goal is to train the action decoder to predict \(A_t\) while forming an intermediate intent state \(I_t\) that represents the local objective served by the action under the instruction.
Standard behavior cloning models \(p_\theta(A_t\mid B_t,q_t)\), leaving this objective implicit.
Let \(I_t\) denote the recovered intent state at the intermediate decoder layer.
At each policy query, \(I_t\) is deterministically recovered from \(B_t\) and \(q_t\), and the decoder models
\begin{equation}
(A_t,V_t,R_t)
\sim
p_\theta(A_t,V_t,R_t\mid B_t,q_t,I_t).
\label{eq:intd_formulation}
\end{equation}
Here, \(A_t\), \(V_t\), and \(R_t\) denote the action, visual-outcome, and textual-purpose supervision targets, while hats denote their decoder predictions.

\subsection{Intent Supervision from Executed Behavior}

Since robot demonstrations do not explicitly label the local objective, we derive intent supervision using a frozen teacher VLM.
For an \(H\)-step demonstrated behavior segment \(\mathcal{W}_t\), let
\(c_A(\mathcal{W}_t)\) denote a coarse textual summary of its demonstrated actions.
Under a functional-intent prompt, a teacher VLM processes
$
    \mathcal{E}_t
    =
    \left(
    o_t,\ell,
    c_A(\mathcal{W}_t),
    \mathrm{vid}_{t:t+H}
    \right)
$
and then generates a functional-purpose statement with token span
\(\mathcal{S}_t\).
We cache the hidden states of both spans from the same autoregressive pass.

\noindent\textbf{Target extraction.}
We extract the intent target from a middle teacher layer to retain both multimodal grounding and semantic abstraction before higher layers specialize toward language generation \citep{kim2026rldx}.
The textual-purpose target uses the final-layer representation of the generated statement, while the visual-outcome target uses a separate frozen encoding of the endpoint observation:
\begin{equation}
\begin{aligned}
I_t^\star
=
\mathrm{pool}_{K_I}\!\left(T_{\mathcal{E}_t}^{(\ell_I)}\right),
\quad
R_t^\star
=
\mathrm{pool}_{K_R}\!\left(T_{\mathcal{S}_t}^{(\ell_R)}\right),
\quad
V_t^\star
=
\mathrm{pool}_{K_V}\!\left(\Phi_{\mathrm{vis}}(o_{t+H})\right).
\end{aligned}
\label{eq:teacher_targets}
\end{equation}
Here \(T_{\mathcal{X}}^{(\ell)}\) denotes teacher hidden states over token span
\(\mathcal{X}\) at layer \(\ell\), and \(\Phi_{\mathrm{vis}}\) is the frozen visual encoder used for endpoint features.
Thus, \(I_t^\star\) retains the multimodal representation formed while the teacher interprets the executed behavior, whereas \(R_t^\star\) provides its linguistic realization.

\noindent\textbf{Visual and textual groundings.}
The endpoint visual target grounds intent in the resulting scene change, while the textual-purpose target grounds it in the functional meaning of the behavior.
These choices build on prior work using predictive visual supervision
\citep{wu2024gr1,black2023susie,hu2025vpp,zhang2025dreamvla}
and language or reasoning traces for robot policies
\citep{zawalski2025ecot,huang2025thinkact,li2025coavla,sumers2023distilling}.

\subsection{Intent-Aware Action Decoding}

To make intent a functional intermediate state rather than an auxiliary readout, the decoder must recover it from the current VLA inputs before final action prediction and make it available to both action and grounding streams. We therefore augment the pretrained flow-matching action decoder with intent queries and visual and textual grounding rows. Let \(Q_t=E_q(q_t)\) denote the proprioceptive-state representation, and let \(Q^I\) denote \(K_I\) clean learnable intent queries.
The decoder processes
\[
    [\,Q_t,\;A_t^\lambda,\;Z_t^{V,\lambda},\;Z_t^{R,\lambda},\;Q^I\,],
\]
where \(A_t^\lambda\) is the noised action input and
\(Z_t^{V,\lambda}\) and \(Z_t^{R,\lambda}\) are self-conditioned visual and
textual grounding inputs.
The clean targets \(I_t^\star\), \(V_t^\star\), and \(R_t^\star\) are used only
for supervision.

\noindent\textbf{Recovering intent.}
We supervise intent at an intermediate layer so that the recovered representation
can shape the remaining decoder computation.
Following intermediate representation alignment in VLA decoders
\citep{zheng2025flare}, we read the contextualized intent-query states
\(H_{I,t}^{L_{\mathrm{tap}}}\), project them into the teacher feature space, and
optimize
\begin{equation}
    \mathcal{L}_I
    =
    \frac{1}{K_I}
    \sum_{k=1}^{K_I}
    \left[
    1-
    \cos
    \left(
    P_I(H_{I,t,k}^{L_{\mathrm{tap}}}),
    \mathrm{sg}(I_{t,k}^\star)
    \right)
    \right],
\label{eq:intent_alignment}
\end{equation}
where \(P_I\) is a projection head and \(\mathrm{sg}\) denotes stop-gradient. The decoder-space states \(H_{I,t}^{L_{\mathrm{tap}}}\) constitute the recovered intent used by the remaining layers.

\noindent\textbf{Action and grounding prediction.}
The action stream retains the flow-matching objective of the pretrained VLA \citep{black2025pi05,bjorck2025grootn1}.
Given noise \(\epsilon_A\) and flow time \(\lambda\), let $
    A_t^\lambda
    =
    (1-\lambda)\epsilon_A+\lambda A_t$ and
    $u_A=A_t-\epsilon_A.$
The decoder predicts \(\hat{u}_A\) using
\begin{equation}
    \mathcal{L}_A
    =
    \mathbb{E}_{\lambda,\epsilon_A}
    \left[
    \left\|
    \hat{u}_A-u_A
    \right\|_2^2
    \right].
\label{eq:action_loss}
\end{equation}

The grounding streams must remain available during decoding without receiving clean grounding targets that are unavailable at deployment.
A no-gradient pass first predicts clean decoder-space grounding states
\(\bar{Z}_t^V\) and \(\bar{Z}_t^R\).
We detach and re-noise these predictions at the sampled flow time:
\[
    Z_t^{V,\lambda}
    =
    (1-\lambda)\epsilon_V
    +
    \lambda\,\mathrm{sg}(\bar{Z}_t^V),
    \qquad
    Z_t^{R,\lambda}
    =
    (1-\lambda)\epsilon_R
    +
    \lambda\,\mathrm{sg}(\bar{Z}_t^R).
\]
Rather than training separate flow generators in each target space, we align the final grounding states with their targets using representation alignment \citep{yu2025repa}. For \(G \in \{V, R\}\),
\begin{equation}
    \mathcal{L}_G
    =
    \mathbb{E}_{\lambda,\epsilon_G}
    \left[
    \frac{1}{K_G}
    \sum_{k=1}^{K_G}
    \left(
    1-
    \cos
    \left(
    P_G(H_{G,t,k}^{L_{\mathrm{out}},\lambda}),
    \mathrm{sg}(G_{t,k}^\star)
    \right)
    \right)
    \right].
\label{eq:grounding_losses}
\end{equation}
Action and grounding rows are jointly processed by the same decoder, allowing the
recovered intent to organize their subsequent prediction.

\noindent\textbf{Intent-dependent information flow.}
To make the recovered intent a functional intermediate state, we organize downstream decoding around it.
All rows are processed jointly, but with asymmetric attention.
Intent queries attend to \(Q_t\), the VLA context \(B_t\), and one another, while remaining isolated from the noised action and grounding rows.
Grounding rows attend to the recovered intent and one another, whereas action rows attend to both intent and grounding rows.
Furthermore, a binary gate \(\alpha\) controls direct \(B_t\) access from non-intent rows.
During training, we sample \(\alpha\sim\mathrm{Bernoulli}(0.5)\), and updates with \(\alpha=0\) require contextual information to pass through the intent queries.

For these intent-mediated updates, we further require downstream prediction to depend on the content of the recovered intent.
We compare decoder continuations under the matched intent states
\(H_{I,t}^{L_{\mathrm{tap}}}\) and batch-shifted states
\(\widetilde{H}_{I,t}^{L_{\mathrm{tap}}}\).
Let \(\mathcal{D}_{\mathrm{right}}\) and \(\mathcal{D}_{\mathrm{swap}}\) denote the corresponding downstream action and grounding losses.
We optimize
\begin{equation}
    \mathcal{L}_{\mathrm{mis}}
    =
    \max
    \left(
    0,\,
    m
    +
    \mathcal{D}_{\mathrm{right}}
    -
    \mathcal{D}_{\mathrm{swap}}
    \right),
\label{eq:intent_mismatch_loss}
\end{equation}
which forces matched intent to yield lower downstream loss than mismatched intent by margin $m$.

\noindent\textbf{Training objective and deployment.}
The full objective is
\begin{equation}
    \mathcal{L}
    =
    \mathcal{L}_A
    +
    \lambda_I\mathcal{L}_I
    +
    \lambda_V\mathcal{L}_V
    +
    \lambda_R\mathcal{L}_R
    +
    \lambda_{\mathrm{mis}}\mathcal{L}_{\mathrm{mis}}.
\label{eq:total_loss}
\end{equation}
At deployment, we set \(\alpha=1\) and remove the teacher, cached targets,
alignment projections, and intent-mismatch branch.
The policy receives the original VLA inputs and internally forms intent and
grounding states together with the action stream.
\begin{table}[htbp]
\centering
\small
\caption{\textbf{Results on SimplerEnv-Bridge.}
Success rates (\%) on the four evaluated tasks.
Controlled variants report mean \(\pm\) sample standard deviation over three evaluation runs, while published baselines retain their originally reported values.
Bold marks the best mean in each column, and underline marks the runner-up.}
\label{tab:simpler-results}
\setlength{\tabcolsep}{5pt}
\resizebox{0.8\textwidth}{!}{%
\begin{tabular}{lccccc}
\toprule
Method & Spoon & Carrot & Stack & EP-Basket & Avg. \\
\midrule
\multicolumn{6}{l}{\emph{General-purpose VLA policies}} \\
Octo-base \citep{octo2024} & 12.5 & 8.3 & 0.0 & 43.1 & 16.0 \\
Octo-small \citep{octo2024} & 47.2 & 9.7 & 4.2 & 56.9 & 29.5 \\
OpenVLA \citep{kim2025openvla} & 0.0 & 0.0 & 0.0 & 4.1 & 1.0 \\
RoboVLMs \citep{liu2024robovlms} & 29.2 & 25.0 & 12.5 & 58.3 & 31.3 \\
SpatialVLA \citep{qu2025spatialvla} & 16.7 & 25.0 & 29.2 & \textbf{100.0} & 42.7 \\
\(\pi_0\) \citep{black2025pi0} & 46.7 & 38.7 & 42.7 & 39.3 & 41.8 \\
\(\pi_0\)-FAST \citep{pertsch2025fast} & 59.0 & 79.0 & \underline{65.0} & 33.0 & 59.0 \\
GR00T-N1.5 \citep{nvidia2025grootn15} & 30.0 & 28.0 & 16.0 & 42.7 & 29.2 \\
\midrule
\multicolumn{6}{l}{\emph{Action abstraction and semantic reasoning}} \\
LAPA \citep{ye2025lapa} & 70.8 & 45.8 & 54.2 & 58.3 & 57.3 \\
UniVLA \citep{bu2025univla} & 52.8 & 55.6 & 2.8 & 80.6 & 47.9 \\
ECoT \citep{zawalski2025ecot,zhang2026scale} & 40.2 & 11.7 & 0.0 & 28.4 & 20.1 \\
\midrule
\multicolumn{6}{l}{\emph{Controlled backbone comparisons}} \\
GR00T-N1.7 \citep{nvidia2026grootn17}
& \secondmeanstd{84.7}{3.1}
& \secondmeanstd{79.3}{9.9}
& \meanstd{57.3}{8.1}
& \meanstd{36.0}{6.0}
& \meanstd{64.3}{3.3} \\
GR00T-N1.7 + future supervision
& \meanstd{81.3}{2.3}
& \meanstd{73.3}{4.2}
& \meanstd{56.7}{3.1}
& \meanstd{60.7}{4.2}
& \secondmeanstd{68.0}{1.0} \\
\rowcolor{green!15}
\textbf{GR00T-N1.7 + \method{} (Ours)}
& \bestmeanstd{88.7}{1.2}
& \bestmeanstd{84.7}{6.1}
& \bestmeanstd{69.3}{5.0}
& \secondmeanstd{96.0}{4.0}
& \bestmeanstd{84.7}{0.8} \\
\(\Delta\) (Ours \(-\) baseline)
& \(+4.0\)
& \(+5.4\)
& \(+12.0\)
& \(+60.0\)
& \(+20.4\) \\
\midrule
\(\pi_{0.5}\) \citep{black2025pi05}
& \meanstd{78.0}{3.5}
& \meanstd{72.7}{7.6}
& \meanstd{32.0}{3.5}
& \meanstd{26.7}{4.2}
& \meanstd{52.3}{2.3} \\
\rowcolor{green!15}
\textbf{\(\pi_{0.5}\) + \method{} (Ours)}
& \meanstd{81.3}{4.2}
& \meanstd{76.0}{4.0}
& \meanstd{39.3}{5.0}
& \meanstd{38.7}{3.1}
& \meanstd{58.8}{3.3} \\
\(\Delta\) (Ours \(-\) baseline)
& \(+3.3\)
& \(+3.3\)
& \(+7.3\)
& \(+12.0\)
& \(+6.5\) \\
\bottomrule
\end{tabular}
}
\vspace{-0.6cm}
\end{table}

\section{Experiments}

We evaluate \method{} across simulation and real-world manipulation settings to answer three questions:
whether behavior-intent supervision improves policy performance across benchmarks and VLA backbones, whether its benefits extend to real-world and out-of-distribution conditions, and whether the recovered latent functions as intent rather than generic additional capacity.
Our evaluation covers SimplerEnv-Bridge and RoboCasa Kitchen in simulation, together with real-world tabletop tasks involving held-out objects and distractors.

\subsection{Experimental Setup}
\label{sec:experiments-setup}

\noindent\textbf{Benchmarks.}
We evaluate \method{} on two simulation benchmarks and a real-world manipulation setting.
SimplerEnv-Bridge~\citep{li2025simpler} covers four tabletop tasks, while RoboCasa Kitchen~\citep{nasiriany2024robocasa} contains 24 household tasks spanning pick-and-place, articulated-object manipulation, and appliance interaction.
Our real-world evaluation contains four tabletop tasks under standard scenes, held-out objects, and distractor conditions.
Dataset, robot-platform, and evaluation-protocol details are provided in \Cref{app:experimental_details}.

\noindent\textbf{Baselines.}
Our primary comparisons use GR00T-N1.7~\citep{nvidia2026grootn17} and \(\pi_{0.5}\)~\citep{black2025pi05}, with each baseline trained using the same demonstrations, optimization budget, and evaluation protocol as its corresponding \method{} model.
We additionally report published benchmark results for context.
On RoboCasa Kitchen, GR00T checkpoints trained with 3,000 demonstrations per task are included only as data-scale references, rather than matched baselines.
For real-world tasks, we compare against the GR00T-N1.7 under the same rollout protocol.
Descriptions and comparison settings for all baselines are provided in \Cref{app:baselines}.

\noindent\textbf{Implementation details.}
We retain each backbone's optimizer, learning rate, and vision-language freezing configuration, and use the same \method{} configuration across benchmarks unless stated otherwise.
By default, we use \(K_I=8\), \(K_R=8\), \(K_V=16\) per camera view, an intent-alignment tap at \(50\%\) of the decoder depth, margin \(m=0.05\), context-dropout rate \(0.5\), and Cosmos-Reason2-8B~\citep{nvidia2025cosmosreason2code} as the teacher VLM.
Full hyperparameters, target-construction details, and implementation choices are provided in \Cref{app:experimental_details}.

\subsection{Results on Simulation Benchmarks}
\label{sec:experiments-simul}

We first test whether behavior-intent supervision improves pretrained VLA policies across distinct simulation benchmarks and backbone architectures.

\noindent\textbf{SimplerEnv-Bridge.}
\Cref{tab:simpler-results} reports success rates on four SimplerEnv-Bridge tasks.
On GR00T-N1.7, \method{} improves average success from \(64.3\%\) to \(84.7\%\), a \(+20.4\) pp gain.
The improvement holds across all four tasks, including tasks where the baseline is already strong.
The largest gain occurs on EP-Basket, where success increases from \(36.0\%\) to \(96.0\%\), while Spoon, Carrot, and Stack improve by \(4.0\), \(5.4\), and \(12.0\) pp, respectively. Excluding EP-Basket, the remaining three tasks improve by \(7.1\) pp on average.
The future-supervision variant reaches \(68.0\%\), while \method{} reaches \(84.7\%\), a further \(+16.7\) pp improvement.
Among the reported results on this task suite, \method{} achieves the highest average success.
The improvement also transfers to \(\pi_{0.5}\), increasing average success from \(52.3\%\) to \(58.8\%\), a \(+6.5\) pp gain, with improvements on every task.

\begin{table}[t]
\centering
\small
\caption{\textbf{Results on RoboCasa Kitchen.}
Success rates (\%) on the 24-task benchmark.
\(G_n\) denotes \(n\) demonstrations per task.
The \(G_{3000}\) checkpoints provide single-value data-scale references, while controlled \(G_{100}\) variants report mean \(\pm\) sample standard deviation over three evaluation runs.
Success rates are averaged within each task category, while Avg. is the macro-average across all 24 tasks.
Bold marks the best mean in each column, and underline marks the runner-up.}
\label{tab:robocasa-results}
\setlength{\tabcolsep}{5pt}
\renewcommand{\arraystretch}{0.95}
\resizebox{0.8\textwidth}{!}{%
\begin{tabularx}{0.92\textwidth}{@{}>{\raggedright\arraybackslash}Xcccc@{}}
\toprule
Method
& \shortstack{Pick-and\\Place}
& \shortstack{Open-or\\Close}
& Others
& Avg. \\
\midrule
GR00T-N1.6 \citep{nvidia2025grootn16} (\(G_{3000}\))
& 43.2
& \underline{81.0}
& 75.8
& 66.2 \\
GR00T-N1.7 \citep{nvidia2026grootn17} (\(G_{3000}\))
& \textbf{53.0}
& 80.8
& \underline{79.0}
& \textbf{70.8} \\
\midrule
GR00T-N1.7 (\(G_{100}\))
& \meanstd{39.4}{1.0}
& \meanstd{75.9}{3.4}
& \meanstd{76.7}{0.9}
& \meanstd{64.1}{1.6} \\
GR00T-N1.7 + future supervision (\(G_{100}\))
& \meanstd{47.0}{4.4}
& \meanstd{80.3}{2.3}
& \meanstd{72.1}{2.0}
& \meanstd{65.8}{0.8} \\
\rowcolor{green!15}
\textbf{GR00T-N1.7 + \method{} (\(G_{100}\), Ours)}
& \secondmeanstd{49.8}{2.3}
& \bestmeanstd{82.8}{1.3}
& \bestmeanstd{79.1}{1.5}
& \secondmeanstd{70.3}{1.7} \\
\(\Delta\) (Ours \(-\) baseline)
& \(+10.4\)
& \(+6.9\)
& \(+2.4\)
& \(+6.2\) \\
\midrule
\(\pi_{0.5}\) \citep{black2025pi05}
& \meanstd{14.0}{1.7}
& \meanstd{55.1}{7.4}
& \meanstd{39.5}{0.9}
& \meanstd{34.9}{2.0} \\
\rowcolor{green!15}
\textbf{\(\pi_{0.5}\) + \method{} (Ours)}
& \meanstd{15.3}{1.4}
& \meanstd{56.1}{3.6}
& \meanstd{53.5}{2.4}
& \meanstd{41.4}{1.4} \\
\(\Delta\) (Ours \(-\) baseline)
& \(+1.3\)
& \(+1.0\)
& \(+14.0\)
& \(+6.5\) \\
\bottomrule
\end{tabularx}
}
\vspace{-8pt}
\end{table}

\begin{wraptable}{r}{0.43\textwidth}
\vspace{-8pt}
\centering
\scriptsize
\caption{\textbf{Reported RoboCasa Kitchen results.}
Average success rates (\%).}
\label{tab:robocasa-published}
\setlength{\tabcolsep}{2.5pt}
\renewcommand{\arraystretch}{1.03}
\begin{tabularx}{\linewidth}{@{}>{\raggedright\arraybackslash}Xcc@{}}
\toprule
Method & Demos/task & Avg. \\
\midrule
\multicolumn{3}{@{}l}{\emph{Base VLA policies}} \\
GR00T-N1 \citep{bjorck2025grootn1}
& 300 & 49.6 \\
\(\pi_0\) \citep{black2025pi0}
& 300 & 62.5 \\
\(\pi_0\)-FAST \citep{pertsch2025fast}
& 300 & 63.6 \\
GR00T-N1.5 \citep{nvidia2025grootn15}
& 300 & 65.7 \\
\midrule
\multicolumn{3}{@{}l}{\emph{Future and video-based methods}} \\
DreamGen \citep{jang2025dreamgen}
& 300 & 57.6 \\
DUST \citep{won2026dust}
& 300 & 58.5 \\
Video Policy \citep{liang2025videopolicy}
& 300 & 66.0 \\
FLARE \citep{zheng2025flare}
& 300 & 66.4 \\
\midrule
\multicolumn{3}{@{}l}{\emph{History and representation methods}} \\
HAMLET \citep{koo2026hamlet}
& 300 & 66.4 \\
RS-CL \citep{kim2026rscl}
& 300 & \underline{69.7} \\
\midrule
\multicolumn{3}{@{}l}{\emph{Behavior-intent supervision}} \\
\rowcolor{green!15}
\textbf{\method{} (Ours)}
& \textbf{100}
& \textbf{70.3} \\
\bottomrule
\end{tabularx}
\vspace{-8pt}
\end{wraptable}

\noindent\textbf{RoboCasa Kitchen.}
\Cref{tab:robocasa-results} reports controlled comparisons on the 24-task RoboCasa Kitchen benchmark.
With GR00T-N1.7 trained on \(100\) demonstrations per task, \method{} improves average success from \(64.1\%\) to \(70.3\%\), with gains across all three task categories.
The future-supervision variant reaches \(65.8\%\), whereas the full method reaches \(70.3\%\), an additional \(+4.5\) pp gain.
The improvement again transfers to \(\pi_{0.5}\), increasing average success from \(34.9\%\) to \(41.4\%\), including a \(+14.0\) pp gain on the Others.

Moreover, \method{} trained with \(100\) demonstrations per task reaches \(70.3\%\) average success, within \(0.5\) pp of the reported GR00T-N1.7 checkpoint trained with \(3{,}000\) demonstrations per task.
\Cref{tab:robocasa-published} further compares against reported RoboCasa Kitchen results from prior methods.
Among these reported results, \method{} achieves the highest average success rates.
Together, the results show that \method{} consistently improves two backbones across different simulation environments.

\subsection{Results on Real-world Benchmarks}

\begin{table}[htbp]
\centering
\small
\caption{\textbf{Results on real-world tasks.}
Success rates (\%) over 50 trials.
ID clean uses training objects without distractors, held-out substitutes unseen objects of the same functional role, and distractors add unrelated items to the scene.
Bold marks the better result within each condition.}
\label{tab:realworld-results}
\setlength{\tabcolsep}{4pt}
\renewcommand{\arraystretch}{0.95}
\resizebox{0.64\textwidth}{!}{%
\begin{tabular}{@{}lcccccc@{}}
\toprule
& \multicolumn{2}{c}{ID clean}
& \multicolumn{2}{c}{Held-out}
& \multicolumn{2}{c}{Distractors} \\
\cmidrule(lr){2-3}\cmidrule(lr){4-5}\cmidrule(lr){6-7}
Task
& Base & \cellcolor{green!15}+ \method{}
& Base & \cellcolor{green!15}+ \method{}
& Base & \cellcolor{green!15}+ \method{} \\
\midrule
Threading
& 92.0 & \cellcolor{green!15}\textbf{96.0}
& \textbf{86.0} & \cellcolor{green!15}84.0
& 70.0 & \cellcolor{green!15}\textbf{82.0} \\
Basket Nesting
& \textbf{94.0} & \cellcolor{green!15}92.0
& 84.0 & \cellcolor{green!15}\textbf{90.0}
& 76.0 & \cellcolor{green!15}\textbf{84.0} \\
Cross-Bin Stacking
& 74.0 & \cellcolor{green!15}\textbf{80.0}
& 62.0 & \cellcolor{green!15}\textbf{72.0}
& 58.0 & \cellcolor{green!15}\textbf{64.0} \\
Drawer Storage
& 24.0 & \cellcolor{green!15}\textbf{36.0}
& 16.0 & \cellcolor{green!15}\textbf{26.0}
& 8.0 & \cellcolor{green!15}\textbf{18.0} \\
\midrule
Average
& 71.0 & \cellcolor{green!15}\textbf{76.0}
& 62.0 & \cellcolor{green!15}\textbf{68.0}
& 53.0 & \cellcolor{green!15}\textbf{62.0} \\
\bottomrule
\end{tabular}%
}
\end{table}

\Cref{tab:realworld-results} reports success rates on the four real-world tasks.
\method{} improves the baseline from \(71.0\%\) to \(76.0\%\) under ID clean, and the gain persists under held-out objects (\(62.0\%\) to \(68.0\%\)) and distractors (\(53.0\%\) to \(62.0\%\)).
Relative retention is comparable under held-out objects (\(87.3\%\) vs.\ \(89.5\%\)) and higher for \method{} under distractors (\(74.6\%\) vs.\ \(81.6\%\)), indicating that the improvement persists as the scene shifts from training conditions.
The gains are concentrated in the longer tasks.
Averaged across the three conditions, Cross-Bin Stacking and Drawer Storage improve by \(7.3\) and \(10.7\) pp, respectively, whereas Threading and Basket Nesting improve by \(4.7\) and \(4.0\) pp.
\Cref{tab:realworld-stages} localizes these differences under ID clean.
On Cross-Bin Stacking, both policies retrieve the base cube at the same rate (\(96.0\%\)) and remain similar after centering (\(90.0\%\) vs.\ \(92.0\%\)), but \method{} retains higher completion when retrieving the top cube (\(82.0\%\) to \(86.0\%\)) and completing the stack (\(74.0\%\) to \(80.0\%\)).
Drawer Storage exhibits a different pattern: the gap is already present at grasping (\(50.0\%\) to \(70.0\%\)) and persists through the subsequent stages.
The two shorter tasks remain near ceiling under ID clean and show only small, mixed differences.

\Needspace{0.1\textheight}
\begin{wraptable}[18]{r}{0.35\textwidth}
\vspace{-8pt}
\centering
\scriptsize
\caption{\textbf{Stage-level completion.}
Fraction of trials (\%) reaching each stage under ID clean (cumulative).}
\label{tab:realworld-stages}
\setlength{\tabcolsep}{3pt}
\renewcommand{\arraystretch}{0.95}
\begin{tabular}{@{}llc>{\columncolor{green!15}}c@{}}
\toprule
Task & Stage & Base & + \method{} \\
\midrule
\multirow{3}{*}{Threading}
& Pole centered & 100.0 & 100.0 \\
& Ring grasped & 96.0 & \textbf{100.0} \\
& Inserted & 92.0 & \textbf{96.0} \\
\midrule
\multirow{2}{*}{Basket Nest.}
& Lifted & \textbf{100.0} & 96.0 \\
& Nested & \textbf{94.0} & 92.0 \\
\midrule
\multirow{4}{*}{Cross-Bin}
& Base retrieved & 96.0 & 96.0 \\
& Base centered & 90.0 & \textbf{92.0} \\
& Top retrieved & 82.0 & \textbf{86.0} \\
& Stacked & 74.0 & \textbf{80.0} \\
\midrule
\multirow{5}{*}{Drawer}
& Grasped & 50.0 & \textbf{70.0} \\
& Transferred & 50.0 & \textbf{70.0} \\
& Opened & 40.0 & \textbf{56.0} \\
& Placed inside & 26.0 & \textbf{38.0} \\
& Closed & 24.0 & \textbf{36.0} \\
\bottomrule
\end{tabular}
\vspace{-8pt}
\end{wraptable}

The stage profiles reveal two different sources of improvement on the longer tasks.
On Cross-Bin Stacking, the policies remain nearly indistinguishable through centering the base cube, and the main gap appears when execution must move from the completed first subgoal to retrieving and stacking the second cube.
This provides the clearest real-world evidence that \method{} improves behavior across a stage transition.
Drawer Storage differs: \method{} already improves grasping and transfer from \(50.0\%\) to \(70.0\%\), and this advantage persists through opening (\(40.0\%\) to \(56.0\%\)), placement (\(26.0\%\) to \(38.0\%\)), and closure (\(24.0\%\) to \(36.0\%\)).
Thus, its gain cannot be attributed solely to a single transition, but instead reflects more reliable execution across the multi-stage sequence.
By contrast, Threading and Basket Nesting have high early-stage completion and little separation under ID clean, leaving less room for improvement.
Together, these results indicate that the benefit of intent supervision becomes more pronounced when successful execution must remain organized across multiple stages, while not requiring every gain to arise from the same failure mode.

\subsection{Analysis and Controlled Studies}
\label{sec:analysis}

We isolate the source of the gain, test whether the recovered state functions as intent, and measure deployment cost.
Unless stated otherwise, controlled analyses use GR00T-N1.7 on a fixed SimplerEnv-Bridge evaluation run shared across all variants.
The main benchmark tables separately report means over three evaluation runs.

\begin{table}[t]
\centering
\scriptsize
\caption{\textbf{Controlled analyses on SimplerEnv-Bridge.}
(a) compares supervision and capacity controls under the same backbone, demonstrations, budget, and evaluation run.
Intent only removes the grounding streams, while future supervision replaces the teacher-intent target with an endpoint visual representation.
(b) reports closed-loop success after replacing the task- or phase-discriminative coordinates of the intent and grounding representations.
All values are success rates (\%).}
\label{tab:intent-controls-main}
\setlength{\tabcolsep}{3pt}
\renewcommand{\arraystretch}{0.95}

\begin{subtable}[t]{0.47\textwidth}
\centering
\resizebox{\linewidth}{!}{%
\begin{tabular}{@{}lccccc@{}}
\toprule
Method & Spoon & Carrot & Stack & EP-Basket & Avg. \\
\midrule
GR00T-N1.7
& 82.0
& 68.0
& 66.0
& 30.0
& 61.5 \\
+ Groundings only
& 88.0
& 82.0
& 38.0
& 32.0
& 60.0 \\
+ Future supervision
& 84.0
& 70.0
& 56.0
& 62.0
& 68.0 \\
+ Free latent
& 78.0
& 76.0
& 48.0
& 26.0
& 57.0 \\
+ Intent only
& 86.0
& 84.0
& 70.0
& 64.0
& 76.0 \\
\rowcolor{green!15}
\textbf{\method{} (Ours)}
& \textbf{90.0}
& \textbf{78.0}
& \textbf{74.0}
& \textbf{100.0}
& \textbf{85.5} \\
\bottomrule
\end{tabular}%
}
\caption{Supervision controls.}
\label{tab:supervision-controls-main}
\end{subtable}
\hfill
\begin{subtable}[t]{0.51\textwidth}
\centering
\resizebox{\linewidth}{!}{%
\begin{tabular}{@{}lcccc|cc|c@{}}
\toprule
& \multicolumn{4}{c}{Injected objective}
& \multicolumn{2}{c}{Forced phase} & \\
\cmidrule(lr){2-5}\cmidrule(lr){6-7}
Task & Spoon & Carrot & Stack & EP-Basket & Early & Late & Noise \\
\midrule
Spoon
& \cellcolor{green!15}\textbf{88.0} & 62.0 & 58.0 & 24.0
& 19.0 & 0.0 & 4.0 \\
Carrot
& 50.0 & \cellcolor{green!15}\textbf{84.0} & 56.0 & 32.0
& 6.0 & 18.0 & 0.0 \\
Stack
& 48.0 & 42.0 & \cellcolor{green!15}\textbf{72.0} & 36.0
& 4.0 & 4.0 & 0.0 \\
EP-Basket
& 42.0 & 48.0 & 44.0 & \cellcolor{green!15}\textbf{94.0}
& 0.0 & 0.0 & 0.0 \\
\midrule
Avg. (diag/off-diag)
& \multicolumn{4}{c|}{84.5 \,/\, 45.2}
& 7.3 & 5.5 & 1.0 \\
\bottomrule
\end{tabular}%
}
\caption{Objective and phase interventions.}
\label{tab:objective-intervention-main}
\end{subtable}
\vspace{-20pt}
\end{table}

\noindent\textbf{Is the gain specific to behavior-intent supervision?}
To distinguish teacher-derived intent from grounding supervision, future visual alignment, and additional latent capacity, we compare controlled variants in \Cref{tab:supervision-controls-main}.
Groundings only and free latent remain below the action-only baseline, while future supervision reaches \(68.0\%\).
Intent supervision alone reaches \(76.0\%\), exceeding the baseline by \(14.5\) pp and future supervision by \(8.0\) pp.
Adding visual and textual groundings further raises success to \(85.5\%\), including a gain from \(64.0\%\) to \(100.0\%\) on EP-Basket.
Thus, teacher-derived intent drives the main improvement, while groundings provide complementary gains on average.
Further details are provided in \Cref{app:supervision_controls}.

\Needspace{0.40\textheight}

\begin{wrapfigure}[31]{r}{0.38\textwidth}
\vspace{-6pt}
\centering
\captionsetup{font=small,skip=3pt}
\captionsetup[subfigure]{font=footnotesize,skip=2pt}

\begin{subfigure}[t]{\linewidth}
\centering
\includegraphics[width=0.92\linewidth]{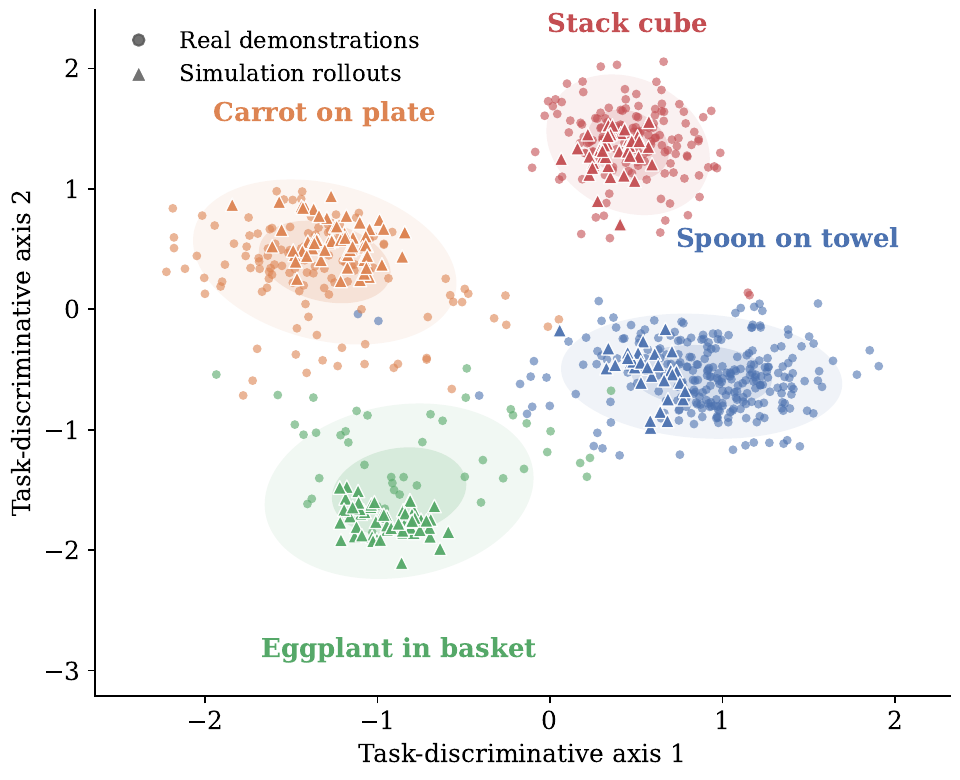}
\caption{\textbf{Objective structure.}
A task-discriminative projection separates recovered intents by behavior objective across real and simulation rollouts.}
\label{fig:intent-objective-main}
\end{subfigure}

\vspace{-2pt}

\begin{subfigure}[t]{\linewidth}
\centering
\includegraphics[width=0.92\linewidth]{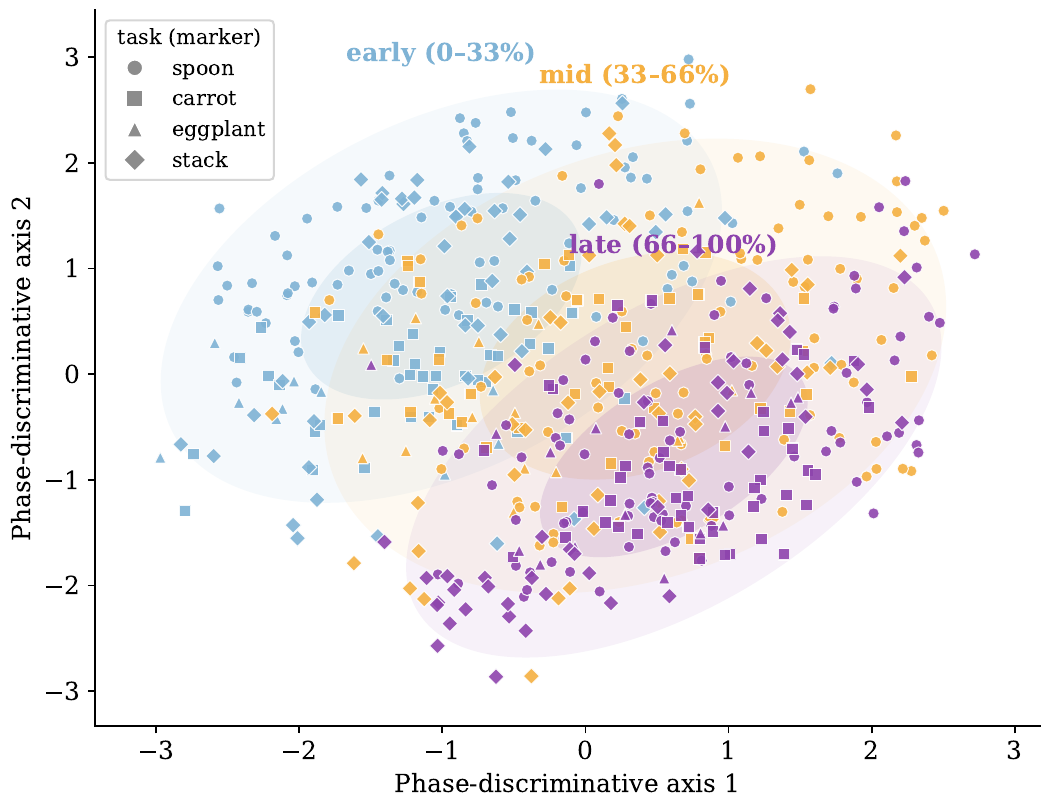}
\caption{\textbf{Progress structure.}
A phase-discriminative projection organizes intents from different tasks by execution stage.}
\label{fig:intent-progress-main}
\end{subfigure}

\vspace{-2pt}

\caption{\textbf{Semantic structure of recovered intent.}
The representation retains both the local objective and progress through the manipulation sequence.}
\label{fig:intent-structure-main}
\vspace{-6pt}
\end{wrapfigure}

\noindent\textbf{Does the recovered representation function as behavioral intent?}
The preceding controls identify intent supervision as the main source of the gain, so we next test whether the recovered state actually functions as intent by examining its objective and stage structure, its use by downstream decoding, and the behavioral effect of changing its content.
As shown in \Cref{fig:intent-structure-main}, recovered intents separate by behavior objective across real demonstrations and simulation rollouts and organize by early, middle, and late execution stages.
Further analyses show that downstream decoding depends on the recovered state, quantify its objective and progress structure, and distinguish it from the particular action sequence used to realize the behavior (\Cref{app:validating_intent,tab:intent-factorization,fig:intent-vs-action-sequence}).

We next test whether this content causally affects execution by editing the task- or phase-discriminative components of the intent and grounding representations.
Same-objective interventions retain \(84.5\%\) success, compared with \(45.2\%\) for cross-objective interventions and \(1.0\%\) under Gaussian corruption (\Cref{tab:objective-intervention-main}).
The uneven cross-objective effects are consistent with recovered-intent task geometry, where Spoon and Carrot form the closest pair and EP-Basket is the most separated (\Cref{fig:task-intent-similarity}).
Phase forcing reduces success to \(7.3\%\) for early content and \(5.5\%\) for late content, while producing stage-consistent behaviors such as approaching without grasping or attempting placement with an empty gripper (\Cref{fig:stage-forced-spoon,fig:stage-forced-carrot,fig:stage-forced-stack}).
Together, these results show that recovered intent represents objective and progress beyond the particular action sequence and causally organizes downstream execution.
Additional protocols and readouts are provided in \Cref{app:validating_intent,app:intent_readout}.

\noindent\textbf{What is the additional computational cost?}
Teacher inference and target construction are performed once offline, and all teacher-side modules are removed at deployment, so runtime overhead comes only from the additional decoder representations.
\method{} increases GR00T-N1.7 from \(3.46\)B to \(3.50\)B parameters and inference time from \(56.1\) to \(61.5\) ms per policy query.
For \(\pi_{0.5}\), the corresponding changes are \(3.62\)B to \(3.64\)B parameters and \(24.4\) to \(29.7\) ms.
Full training-time and resource measurements are reported in \Cref{app:computational_resources,tab:computational_cost}.
\vspace{-5pt}

\section{Conclusion}
\label{sec:conclusion}
\vspace{-5pt}

We presented \method{}, a framework for distilling behavior-level intent into pretrained VLA action decoders.
While behavior cloning supervises executed actions and future-based objectives supervise particular realizations, neither directly identifies the local objective served by the behavior.
During training, a frozen teacher VLM interprets demonstrated behavior, and the decoder learns to recover the resulting intent from standard VLA inputs and use it to organize action and grounding prediction.
Across SimplerEnv-Bridge, RoboCasa Kitchen, and real-world manipulation tasks, \method{} improves strong VLA backbones without requiring the teacher at deployment.
Controlled studies further show that the gains arise specifically from intent supervision and that the recovered state represents objective and progress, is used by the policy, and influences closed-loop execution.
These results position behavior intent as a compact intermediate representation that shifts VLA learning from reproducing executions toward understanding what each behavior should accomplish.

% \section{Reproducibility Statement}

% Appendix~\cref{app:experimental_details} documents the backbone integration, datasets, real-world platform, teacher prompt and target construction, implementation details, optimization settings, checkpoint selection, and computational resources.
% Appendix~\ref{app:additional_ablations} provides the exact configurations and protocols for the controlled ablations, representation analyses, and closed-loop interventions, while Appendix~\ref{app:additional_results} reports the full per-task results.
% Anonymous code, configuration files, target-construction scripts, and evaluation assets are provided in the supplementary material.

\bibliographystyle{iclr2027_conference}
\bibliography{references}

\appendix
% A_appendix_variants documents old design alternatives from the HID-era draft
% (contrastive forcing, vision foresight head) — no longer part of the method,
% kept on disk only.
% \input{sections/A_appendix_variants}
\appendix

%%%%%%%%%%%%%%%%%%%%%%%%%%%%%%%%%%%%%%%%%%%%%%%%%%%%%%%%%%%%%%%%%%%%%%%%%%%%%%%%
%%%%%%%%%%%%%%%%%%%%%%%%%%%%%%%%%%%%%%%%%%%%%%%%%%%%%%%%%%%%%%%%%%%%%%%%%%%%%%%%

\section{Experimental Details}
\label{app:experimental_details}

\begin{figure}[htbp]
\centering
\includegraphics[width=\linewidth]{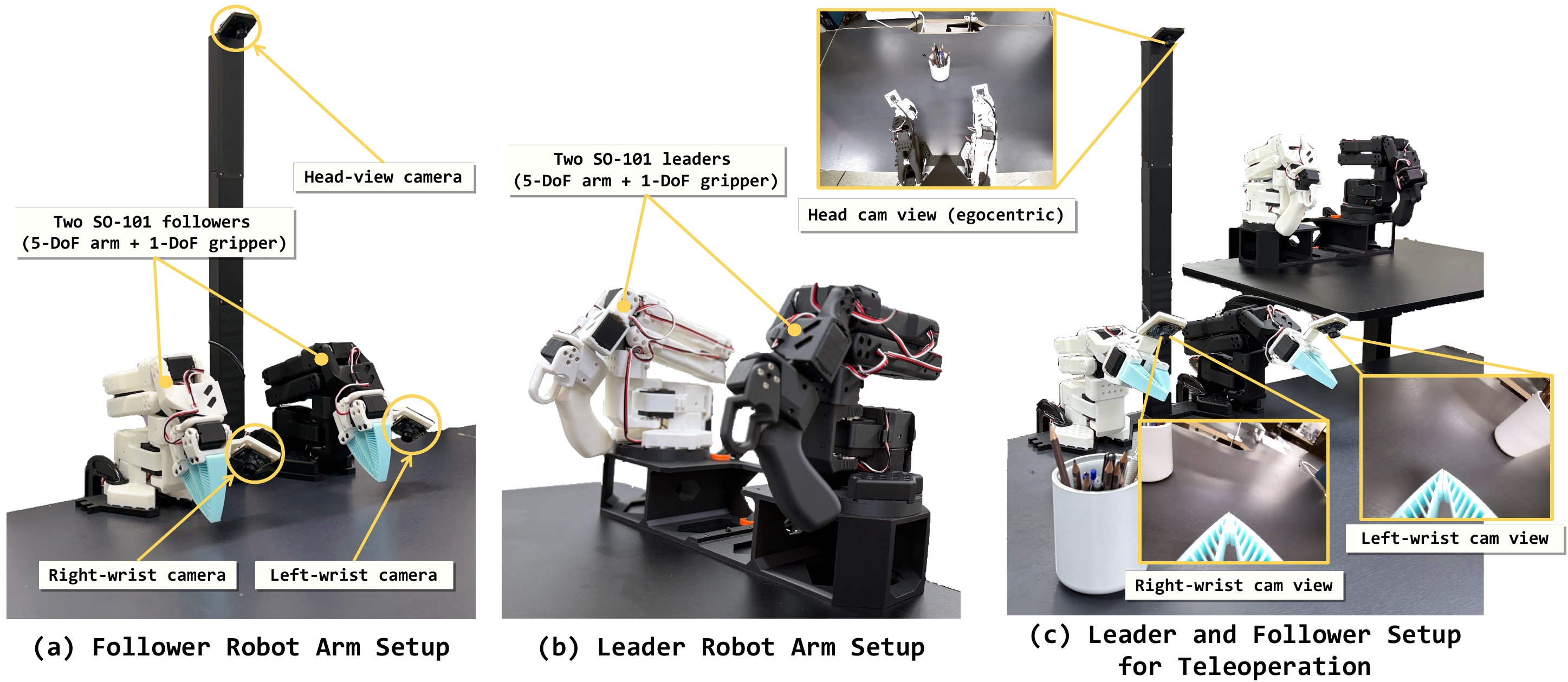}
\caption{\textbf{Real-robot teleoperation and evaluation platform.}
We specify the leader and follower robot arm setups, consisting of 5-DoF arms with 1-DoF grippers, along with the multi-camera views.}
\label{fig:robot_setting}
\end{figure}

%%%%%%%%%%%%%%%%%%%%%%%%%%%%%%%%%%%%%%%%
\subsection{Model Details}
\label{app:model_details}

\noindent\textbf{Backbone VLAs.}
We instantiate \method{} on two flow-matching VLA backbones, GR00T-N1.7-3B \citep{nvidia2026grootn17} and \(\pi_{0.5}\) \citep{black2025pi05,physical_intelligence_openpi}, using their official implementations.
For both models, we freeze the pretrained vision-language module and fine-tune the action decoder together with the newly introduced \method{} parameters.

\noindent\textbf{\method{} integration.}
For GR00T-N1.7, we append self-conditioned visual and textual grounding rows and clean intent queries \(Q^I\) to the pretrained DiT token stream.
The grounding rows follow the backbone's time-conditioned normalization, while the intent queries use standard layer normalization and are never noised.
The recovered intent is supervised at block 16 of 32.
For \(\pi_{0.5}\), the action-expert suffix is organized as
\[
[Q_t,A_t^\lambda,Z_t^{V,\lambda},Z_t^{R,\lambda},Q^I],
\]
where \(Q_t\) is the proprioceptive-state representation, \(A_t^\lambda\) is the noised action sequence, \(Z_t^{V,\lambda}\) and \(Z_t^{R,\lambda}\) are self-conditioned visual and textual grounding inputs, and \(Q^I\) contains the clean intent queries.
The recovered intent is supervised at layer 9 of 18.
We apply the same asymmetric information-flow pattern during training and sampling.
Intent queries attend to the frozen VLA context, proprioceptive state, and one another while remaining isolated from the noised action and grounding rows.
Visual and textual grounding rows attend to the recovered intent and one another, whereas action rows attend to the intent together with both grounding streams.
For \(\pi_{0.5}\), we cache the frozen prefix keys and values during training.

\noindent\textbf{Trainable and frozen components.}
The vision-language modules, including their visual encoders, remain frozen for both backbones.
We optimize the complete action decoder together with the added intent and grounding parameters.
These parameters include intent-query embeddings, positional and camera-view embeddings, and three-layer SiLU MLPs for encoding and projecting the target representations.
The visual target has width \(d_V=2048\), the teacher targets have width \(d_T=4096\), and the added representations are mapped to the backbone-specific decoder width \(d_D\).
Checkpoint-tensor counting gives 3.455B parameters for the GR00T-N1.7 baseline and 3.502B for GR00T-N1.7+\method{}, corresponding to 46.4M additional parameters or a \(1.3\%\) increase.
For \(\pi_{0.5}\), the corresponding counts are 3.617B and 3.640B, giving 23.1M additional parameters or a \(0.64\%\) increase.

\noindent\textbf{Model dimensions.}
Unless stated otherwise, we use \(K_I=8\) intent queries, \(K_R=8\) textual grounding rows, and \(K_V=16\) visual grounding rows per camera.
Bridge uses one camera and therefore 16 visual rows, while RoboCasa Kitchen uses three cameras and therefore 48 visual rows.
We use an intent-mismatch margin of \(m=0.05\) and a context-dropout probability of \(0.5\).

%%%%%%%%%%%%%%%%%%%%%%%%%%%%%%%%%%%%%%%%
\subsection{Datasets}
\label{app:datasets_protocols}

\noindent\textbf{SimplerEnv-Bridge.}
We evaluate on SimplerEnv-Bridge \citep{li2025simpler}, a simulation benchmark for evaluating real-world manipulation policies trained on BridgeData V2 \citep{walke2023bridgedata}.
We use the LeRobot conversion of BridgeData V2, which contains 53,192 episodes and 1,893,026 frames recorded at 5 Hz.
Each sample contains a primary RGB observation at \(256 \times 256\) resolution, an 8-dimensional robot state, and a 7-dimensional action.
We evaluate four WidowX manipulation tasks: spoon on towel, carrot on plate, stack cube, and put eggplant in basket.
Our controlled GR00T-N1.7 and \(\pi_{0.5}\) comparisons use three independent evaluation runs with 50 episodes per task in each run.
We report the mean and sample standard deviation of the three run-level success rates.
Published baseline results follow their original evaluation protocols.

\noindent\textbf{RoboCasa Kitchen.}
RoboCasa Kitchen \citep{nasiriany2024robocasa} is a simulation benchmark for household manipulation in diverse kitchen environments.
We use its machine-generated dataset with 100 demonstrations for each of 24 tasks, giving 2,400 episodes and 689,595 frames recorded at 20 Hz.
Each sample contains left, right, and wrist RGB observations at \(256 \times 256\) resolution, a 53-dimensional robot state, and a 12-dimensional action.
The benchmark contains 8 pick-and-place tasks, 6 open-or-close tasks, and 10 additional appliance and interaction tasks.
We use three independent evaluation runs with 50 episodes per task and run for both GR00T-N1.7 and \(\pi_{0.5}\).
Category and overall averages are computed within each run and then summarized by their mean and sample standard deviation across the three runs.

\begin{figure}[htbp]
\centering
\includegraphics[width=\linewidth]{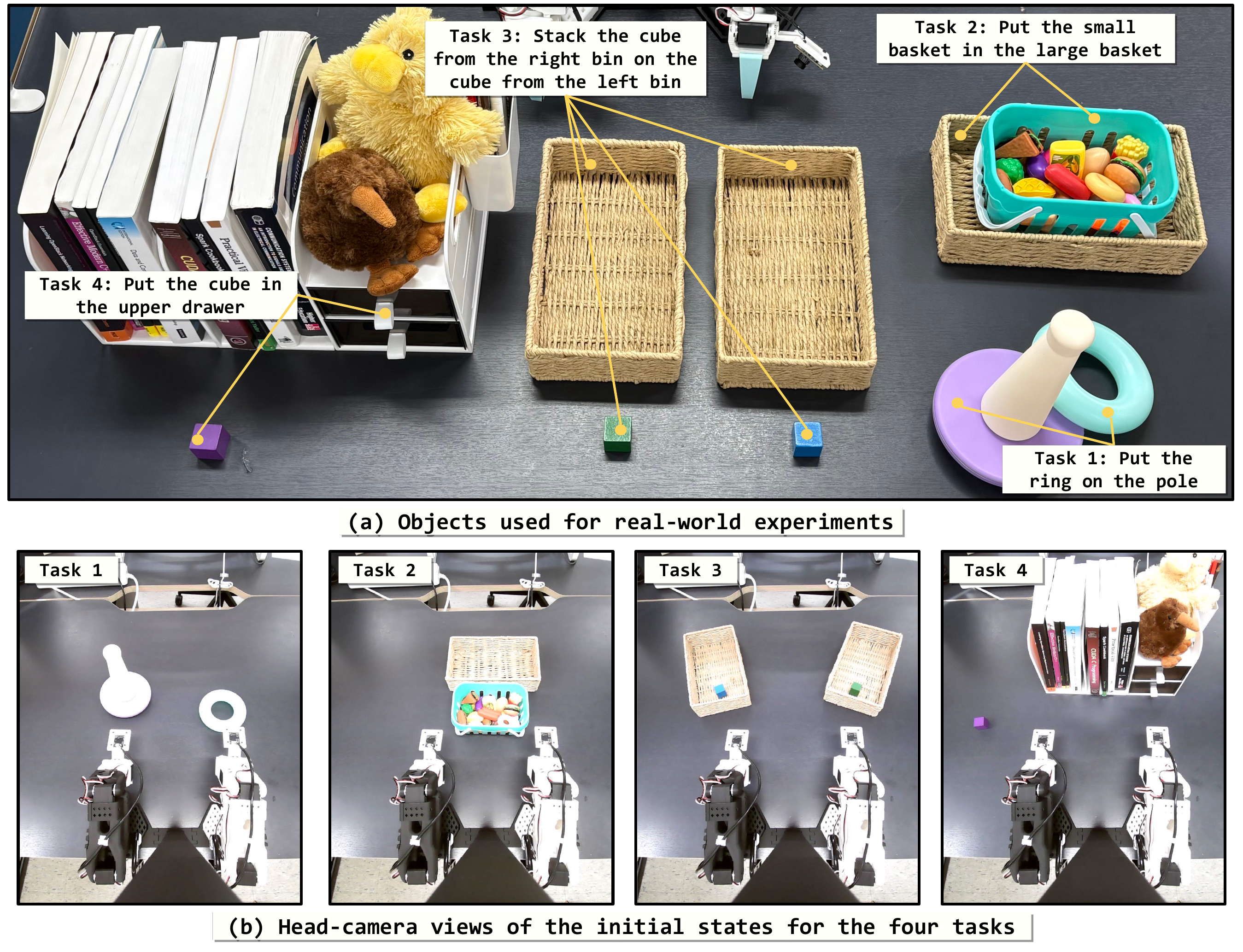}
%\fbox{\begin{minipage}[c][0.28\linewidth][c]{0.98\linewidth}
%\centering
%textcolor{gray}{\texttt{[Figure placeholder: real-world task suite]}}
%\end{minipage}}
\caption{\textbf{Real-world task suite.}
Four tabletop tasks used for real-world experiments.}
\label{fig:real_world_tasks}
\end{figure}

\noindent\textbf{Real-world platform.}
As shown in \Cref{fig:robot_setting}, we use a teleoperation setup consisting of SO-101 robot arms for both the leader and follower, equipped with 1-DoF grippers.
Three camera views are provided: an egocentric view mounted on the head structure and two local views on the wrists.

On this platform, we design four tabletop tasks as illustrated in \Cref{fig:real_world_tasks}, each targeting a distinct capability and ordered by horizon length. For each task, we collect 100 demonstrations and train a single multitask policy per method on identical data.

\begin{itemize}
    \item \emph{Threading} (bimanual) has the left arm bring a pole to the center, after which the right arm inserts a specified ring onto it, testing precise alignment.
    \item \emph{Basket Nesting} (bimanual) lifts a small basket with both arms into a larger one, testing simultaneous cooperation on a shared object that a single arm cannot lift.
    \item \emph{Cross-Bin Stacking} (bimanual) has the left arm place a cube from the left bin at the center, after which the right arm stacks a cube from the right bin on top.
    \item \emph{Drawer Storage} (bimanual) has the left arm pass an object to the right arm, which then opens the specified drawer of a two-tier unit, places the object inside, and closes it.
\end{itemize}

%%%%%%%%%%%%%%%%%%%%%%%%%%%%%%%%%%%%%%%%
\subsection{Baselines}
\label{app:baselines}

We briefly describe the baseline methods included in our evaluations.

\begin{itemize}
\setlength{\itemsep}{2pt}

\item \textbf{Octo} \citep{octo2024} is a transformer-based generalist policy trained on cross-embodiment robot data with a diffusion-based action head.

\item \textbf{OpenVLA} \citep{kim2025openvla} fine-tunes a pretrained vision-language model to autoregressively predict discretized action tokens.

\item \textbf{RoboVLMs} \citep{liu2024robovlms} studies the integration of pretrained vision-language models with continuous and discrete action-prediction architectures.

\item \textbf{SpatialVLA} \citep{qu2025spatialvla} incorporates explicit 3D spatial information through Ego3D positional encoding and adaptive action grids.

\item \textbf{\(\pi_0\)} \citep{black2025pi0} combines a pretrained vision-language prefix with a flow-matching action expert for continuous action generation.

\item \textbf{\(\pi_0\)-FAST} \citep{pertsch2025fast} predicts actions autoregressively using frequency-space action tokenization.

\item \textbf{\(\pi_{0.5}\)} \citep{black2025pi05} extends the \(\pi_0\) family toward open-world generalization and serves as our second backbone.

\item \textbf{GR00T-N1} \citep{bjorck2025grootn1} combines a vision-language module with a flow-matching diffusion-transformer action decoder.

\item \textbf{GR00T-N1.5} \citep{nvidia2025grootn15} improves the architecture, training data, and post-training procedure of GR00T-N1.

\item \textbf{GR00T-N1.6} \citep{nvidia2025grootn16} further updates the GR00T pretraining mixture and post-training recipe.

\item \textbf{GR00T-N1.7} \citep{nvidia2026grootn17} is our primary backbone.

\item \textbf{LAPA} \citep{ye2025lapa} learns discrete latent actions between video frames and pretrains a vision-language model to predict these representations before robot-action fine-tuning.

\item \textbf{UniVLA} \citep{bu2025univla} learns task-centric latent actions from cross-embodiment videos and decodes the predicted latent actions into embodiment-specific robot trajectories.

\item \textbf{ECoT} \citep{zawalski2025ecot} generates embodied reasoning traces before predicting robot actions.

\item \textbf{DreamGen} \citep{jang2025dreamgen} generates synthetic robot trajectories with a video world model and recovers corresponding pseudo-actions.

\item \textbf{DUST} \citep{won2026dust} jointly predicts future observations and actions using interacting diffusion streams with decoupled flow-matching objectives.

\item \textbf{Video Policy} \citep{liang2025videopolicy} combines behavior-video generation and action prediction within an end-to-end robot policy.

\item \textbf{FLARE} \citep{zheng2025flare} aligns intermediate action-decoder representations with latent features of future observations.

\item \textbf{HAMLET} \citep{koo2026hamlet} incorporates observation history through moment tokens and a temporal memory module.

\item \textbf{RS-CL} \citep{kim2026rscl} regularizes vision-language representations with a contrastive objective supervised by distances between robot proprioceptive states.

\end{itemize}

Our controlled comparisons use GR00T-N1.7 and \(\pi_{0.5}\) baselines trained with the same demonstrations, optimization budgets, and evaluation protocols as their corresponding \method{} models. Moreover, future supervision replaces the teacher-intent target with FLARE-style alignment of mid-depth decoder representations to latent features of the future observation, under the same backbone and training budget.
The exact configurations of the supervision and capacity controls are provided in \Cref{app:supervision_controls}.

%%%%%%%%%%%%%%%%%%%%%%%%%%%%%%%%%%%%%%%%
\subsection{Target Construction}
\label{app:target_construction}

\noindent\textbf{Teacher inputs and prompt.}
We use the frozen Cosmos-Reason2-8B model \citep{nvidia2025cosmosreason2code} as the teacher.
For each \(H\)-step demonstrated behavior segment \(\mathcal W_t\), the teacher receives the current observation, language instruction, a coarse textual action summary \(c_A(\mathcal W_t)\), and the corresponding execution video.
In a single autoregressive generation pass, the teacher first processes this multimodal evidence and then generates a functional-purpose statement describing what the executed behavior accomplishes under the instruction.
The prompt asks the teacher to reason in at most four sentences about the state change occurring during the segment and how it advances the instruction.
It then requests exactly one sentence of approximately 30 to 50 tokens that describes the object-level state change or immediate local subgoal.
Low-level control values, camera motion, references to the video, and additional labels or explanations are explicitly excluded.
We use the same prompt template for Bridge and RoboCasa Kitchen.

\begin{tcolorbox}[breakable,title={Prompt for Teacher Target Construction},colback=gray!5,colframe=black!70,sharp corners,boxrule=0.4pt,left=1.5mm,right=1.5mm,top=1.5mm,bottom=1.5mm,fonttitle=\bfseries\small]
\small
\noindent\textbf{System Prompt}

You are a robot manipulation reasoning model.
You will see a short video of an approximately 1.6-second segment from a longer robot task, a short multi-step segment that may include contact, object motion, or clear setup for the next contact, plus a coarse summary of the commanded robot motion.

First write a brief reasoning block inside \texttt{<think>...</think>}, using at most four sentences.
In the reasoning, infer what changes over this segment: which object or relation is acted on, what state transition occurs, and why this segment advances the episode instruction.

Immediately after \texttt{</think>}, write exactly one sentence of roughly 30--50 tokens.
This sentence is the segment's functional intent: the object-level outcome that the behavior accomplishes or clearly progresses toward.
Describe the intended object-level state change, not low-level motion, camera appearance, or the video itself.
State only changes that are visible or strongly implied by the segment.
If contact or object motion has not happened yet, describe the immediate setup or alignment achieved by the segment.

Do not mention ``the video,'' ``the clip,'' ``frames,'' ``I,'' or ``the robot intends.''
Do not add a label, bullet, prefix, or explanation after \texttt{</think>}.
Output only the one sentence after \texttt{</think>}.

\vspace{0.3cm}
\noindent\textbf{User Prompt}

Episode instruction: \texttt{\{instruction\}}

Coarse commanded motion over this segment: \texttt{\{action\_summary\}}

The video frames show an approximately 1.6-second multi-step segment with \texttt{\{n\}} frames.

Identify the functional intent of this behavior segment by stating the object-level outcome it accomplishes or clearly progresses toward under the episode instruction.
\end{tcolorbox}

\noindent\textbf{Action and video preprocessing.}
We convert each continuous action segment into a compact textual representation before providing it to the teacher.
For each action dimension, we compute dataset-specific percentile boundaries from the training set and map the continuous values to coarse discrete motion bins.
The resulting sequence of bucketed values is serialized as a short description of the translational, rotational, and gripper behavior over the segment.
The gripper dimension is interpreted using the open and close convention of each dataset rather than assuming a shared sign convention across embodiments.
This representation provides the teacher with the overall motion pattern while avoiding long sequences of raw floating-point values.
The execution video covers the same temporal interval as the summarized action segment.
On Bridge, we use the 8 frames corresponding to an 8-step segment recorded at 5 Hz.
On RoboCasa Kitchen, each target covers a 32-step segment recorded at 20 Hz, from which we uniformly sample 16 frames.
Both therefore represent an approximately 1.6-second behavior window.
The current observation, action segment, and video clip are matched using their episode identifier and segment start index, ensuring that all teacher inputs describe the same executed behavior.

\noindent\textbf{Intent and textual-purpose targets.}
Let \(\mathcal E_t\) denote the teacher-input span containing the current observation, instruction, coarse action summary, and execution video.
Let \(\mathcal S_t\) denote only the final functional-purpose statement generated after the reasoning span.
During the same autoregressive pass, we cache the hidden states produced while processing \(\mathcal E_t\) and while generating \(\mathcal S_t\).
We construct the intent target \(I_t^\star\) from the layer-18 hidden states over \(\mathcal E_t\), corresponding to the midpoint of the teacher.
Given the input-span hidden states \(H_{\mathcal E_t}^{(18)}\), we partition the token sequence into \(K_I=8\) approximately equal contiguous regions and average the hidden states within each region:
\[
I_{t,k}^\star
=
\frac{1}{|\mathcal G_{I,k}|}
\sum_{i\in\mathcal G_{I,k}}
H_{\mathcal E_t,i}^{(18)},
\qquad k=1,\ldots,K_I.
\]
This preserves the coarse ordering of the multimodal teacher-input representation while producing a fixed number of target slots.
We construct the textual-purpose target \(R_t^\star\) from the final-layer hidden states over \(\mathcal S_t\).
The statement tokens are likewise partitioned into \(K_R=8\) contiguous regions and mean-pooled:
\[
R_{t,k}^\star
=
\frac{1}{|\mathcal G_{R,k}|}
\sum_{i\in\mathcal G_{R,k}}
H_{\mathcal S_t,i}^{(\mathrm{final})},
\qquad k=1,\ldots,K_R.
\]
Both targets have hidden width \(d_T=4096\).
Thus, \(I_t^\star\) is the evidence-side multimodal representation formed while the teacher interprets the demonstrated behavior, whereas \(R_t^\star\) captures its final linguistic realization.

\noindent\textbf{Generated-response intent target for the target-source ablation.}
We additionally construct an alternative intent target from the teacher-generated response.
Let \(\mathcal Y_t\) denote the complete response span generated after processing \(\mathcal E_t\), including both the reasoning block and the final functional-purpose statement.
Using the same teacher layer as the default intent target, we partition the hidden states over \(\mathcal Y_t\) into \(K_I=8\) approximately equal contiguous regions and mean-pool each region:
\[
I_{t,k}^{\star,\mathrm{resp}}
=
\frac{1}{|\mathcal G_{\mathrm{resp},k}|}
\sum_{i\in\mathcal G_{\mathrm{resp},k}}
H_{\mathcal Y_t,i}^{(18)},
\qquad k=1,\ldots,K_I.
\]
This alternative preserves the same target width and number of slots as the default evidence-side target, while replacing the representation formed over the multimodal evidence with one formed over the teacher's complete generated interpretation.

\noindent\textbf{Visual-outcome target.}
We construct the visual-outcome target \(V_t^\star\) from the observation at the endpoint of the behavior segment.
Specifically, the endpoint is \(o_{t+8}\) for Bridge and \(o_{t+32}\) for RoboCasa Kitchen, matching the temporal window used for the corresponding teacher target.
We encode this observation using the frozen visual encoder of Cosmos-Reason2-2B.
The encoder produces 64 regional features with hidden width \(d_V=2048\), which we reduce to \(K_V=16\) ordered target slots using fixed 64-to-16 spatial pooling.
For multi-camera observations, each view is encoded and pooled independently.
Bridge uses one camera and therefore produces 16 visual target slots.
RoboCasa Kitchen uses left, right, and wrist cameras and therefore produces 48 slots in total.
Samples that do not contain a valid endpoint observation near an episode boundary are excluded from the visual-alignment loss.
All teacher and visual targets are generated once offline and reused across policy-training runs.

%%%%%%%%%%%%%%%%%%%%%%%%%%%%%%%%%%%%%%%%
\subsection{Implementation}
\label{app:implementation}

\noindent\textbf{Decoder augmentation.}
We preserve the native state and action representations of each backbone and append self-conditioned visual grounding rows \(Z_t^{V,\lambda}\), self-conditioned textual grounding rows \(Z_t^{R,\lambda}\), and clean intent queries \(Q^I\) to the action-decoder sequence.
For GR00T-N1.7, the added rows are incorporated into the diffusion-transformer stream.
For \(\pi_{0.5}\), they are appended to the action-expert suffix following the frozen vision-language prefix.
All added rows use learned positional embeddings, and visual rows additionally use camera-view embeddings.
For \(\pi_{0.5}\), we cache the frozen prefix keys and values during training.

\noindent\textbf{Asymmetric information flow.}
Intent queries attend to the robot state, frozen VLA context, and one another, but not to the noised action or grounding rows.
Visual and textual grounding rows attend to the recovered intent and one another, while action rows attend to the recovered intent and both grounding streams.
Direct context access from action and grounding rows is controlled by a binary gate \(\alpha\), whereas intent queries always retain access to the VLA context.
During training, we sample \(\alpha\sim\mathrm{Bernoulli}(0.5)\), so updates with \(\alpha=0\) require the downstream rows to obtain contextual information through the intent queries.
For \(\pi_{0.5}\), this gate is applied to the image and language prefix tokens, while access to the state representation remains available.
We use the same information-flow structure during training and sampling and set \(\alpha=1\) at deployment.

\noindent\textbf{Intent alignment.}
We extract the contextualized intent-query states \(H_{I,t}^{L_{\mathrm{tap}}}\) at the midpoint of the decoder, using block 16 for GR00T-N1.7 and layer 9 for \(\pi_{0.5}\).
A three-layer SiLU MLP projects each intent state to the teacher feature space.
The projected rows are aligned one-to-one with the eight teacher intent targets \(I_t^\star\) using cosine distance, with stop-gradient applied to the targets.
The decoder-space states \(H_{I,t}^{L_{\mathrm{tap}}}\), rather than their projected teacher-space representations, are used as the recovered intent by the remaining decoder layers.

\noindent\textbf{Intent-mismatch training.}
For updates with \(\alpha=0\), we construct mismatched intent states by cyclically shifting the recovered intent representations within the batch.
Starting from the intent-alignment layer, the remaining decoder blocks are evaluated once with the original intent and once with the mismatched intent.
We optimize
\[
\mathcal L_{\mathrm{mis}}
=
\max\left(0,m+\mathcal D_{\mathrm{right}}-\mathcal D_{\mathrm{swap}}\right),
\]
where \(\mathcal D_{\mathrm{right}}\) and \(\mathcal D_{\mathrm{swap}}\) denote the combined downstream action and grounding losses, and \(m=0.05\).
The mismatched donor representations are detached so that gradients do not propagate through the donor examples.

\noindent\textbf{Grounding self-conditioning and deployment.}
Clean teacher targets are used only as supervision and are never supplied as decoder inputs.
A no-gradient forward pass first estimates clean decoder-space visual and textual grounding states.
These estimates are detached, re-noised at the sampled flow time, and used as grounding inputs to the gradient pass.
During inference, the visual and textual states are carried across denoising steps using the same self-conditioning convention.
The teacher VLM, target representations, alignment projections, and intent-mismatch branch are used only during training.
The deployed policy receives the original VLA inputs and internally forms intent and grounding states together with the action stream.

%%%%%%%%%%%%%%%%%%%%%%%%%%%%%%%%%%%%%%%%
\subsection{Training Details}
\label{app:training_details}

\noindent\textbf{Optimization and learning-rate schedules.}
We follow the official fine-tuning implementations of GR00T-N1.7 \citep{nvidia2026grootn17} and \(\pi_{0.5}\) \citep{black2025pi05,physical_intelligence_openpi}.
For each backbone, the baseline and \method{} variant use the same benchmark-specific training budget.
For GR00T-N1.7, we use AdamW with a learning rate of \(10^{-4}\).
We train for 60,000 steps with a global batch size of 480 on RoboCasa Kitchen and for 20,000 steps with a global batch size of 1,536 on Bridge.
For \(\pi_{0.5}\), we use AdamW with a maximum gradient norm of 1.0 and linearly warm up the learning rate for 1,000 steps to \(5\times10^{-5}\), which is held constant for the remainder of training.
We use the same benchmark-specific batch sizes and training steps as in the corresponding GR00T-N1.7 experiments.
All models are trained with bfloat16 computation, while loss values are accumulated in float32.
The global batch sizes are distributed across three GPUs without gradient accumulation.

\noindent\textbf{Loss configuration.}
We set \(\lambda_I=\lambda_V=\lambda_R=0.5\), \(\lambda_{\mathrm{mis}}=0.1\), and the intent-mismatch margin to \(m=0.05\).
The weights of the visual and textual grounding losses are linearly increased from zero to their final values during the first 1,000 training steps.
The action and intent-alignment losses are active from the beginning of training.
We use a context-dropout probability of 0.5.

\noindent\textbf{Backbone-specific training configurations.}
For both backbones, we freeze the pretrained vision-language module and optimize the action decoder together with the newly introduced \method{} parameters.
The teacher targets are defined over an approximately 1.6-second behavior window, corresponding to \(H=8\) steps on Bridge and \(H=32\) steps on RoboCasa Kitchen.
For GR00T-N1.7, the decoder retains its native 40-step output sequence, while the action and auxiliary losses are applied to the benchmark-valid segment of 8 steps on Bridge and 32 steps on RoboCasa Kitchen.
For \(\pi_{0.5}\), we retain its native 10-step action horizon and mask the action loss to the valid action dimensions, excluding dimensions introduced only by zero padding.
The teacher behavior window is defined independently of the backbone-native policy horizon.
We cache the keys and values of the frozen \(\pi_{0.5}\) prefix during training.

\noindent\textbf{Checkpoint selection.}
On RoboCasa Kitchen, we evaluate the final checkpoint obtained after 60,000 training steps.
On Bridge, we evaluate checkpoints at 5,000, 10,000, 15,000, and 20,000 steps using two validation rollout seeds and apply the same selection procedure to both backbones.

\noindent\textbf{Controlled ablation protocol.}
The mechanism, policy-level teacher, and intent-target-source ablations use GR00T-N1.7 on SimplerEnv-Bridge.
Unless stated otherwise, all variants use the same demonstrations, student architecture, optimization budget, checkpoint-selection procedure, and fixed evaluation run as the controlled analyses in \Cref{sec:analysis}.
Only the component or target source identified by each ablation is changed.

%%%%%%%%%%%%%%%%%%%%%%%%%%%%%%%%%%%%%%%%
\subsection{Computational Resources}
\label{app:computational_resources}

All experiments are conducted on a single node with three NVIDIA B200 GPUs, each with 183 GB of memory.
Teacher targets are generated once offline and reused across all subsequent policy-training runs.
We summarize parameter counts, inference time per policy query, and benchmark-specific training time in \Cref{tab:computational_cost}.
All inference-time measurements are obtained on a single NVIDIA B200 GPU.

\begin{table}[htbp]
\centering
\small
\caption{\textbf{Modification cost across VLA backbones.}
We report checkpoint-tensor parameter counts, inference time per policy query, and end-to-end training time.
Inference is measured on a single NVIDIA B200 GPU, while training uses three NVIDIA B200 GPUs.}
\label{tab:computational_cost}
\setlength{\tabcolsep}{4pt}
\begin{tabular}{lcccc}
\toprule
Method & Parameters & Inference time (ms) & Bridge training (h) & RoboCasa training (h) \\
\midrule
GR00T-N1.7
& 3.455B
& 56.1 \((1.00\times)\)
& \(\sim\)17
& \(\sim\)17 \\
\rowcolor{green!15}
+\method{}
& 3.502B \((+46.4\mathrm{M},+1.3\%)\)
& 61.5 \((1.10\times)\)
& \(\sim\)19
& \(\sim\)19 \\
\midrule
\(\pi_{0.5}\)
& 3.617B
& 24.4 \((1.00\times)\)
& \(\sim\)23
& \(\sim\)23 \\
\rowcolor{green!15}
+\method{}
& 3.640B \((+23.1\mathrm{M},+0.64\%)\)
& 29.7 \((1.21\times)\)
& \(\sim\)29
& \(\sim\)26 \\
\bottomrule
\end{tabular}
\end{table}

%%%%%%%%%%%%%%%%%%%%%%%%%%%%%%%%%%%%%%%%%%%%%%%%%%%%%%%%%%%%%%%%%%%%%%%%%%%%%%%%
%%%%%%%%%%%%%%%%%%%%%%%%%%%%%%%%%%%%%%%%%%%%%%%%%%%%%%%%%%%%%%%%%%%%%%%%%%%%%%%%
%%%%%%%%%%%%%%%%%%%%%%%%%%%%%%%%%%%%%%%%%%%%%%%%%%%%%%%%%%%%%%%%%%%%%%%%%%%%%%%%
%%%%%%%%%%%%%%%%%%%%%%%%%%%%%%%%%%%%%%%%%%%%%%%%%%%%%%%%%%%%%%%%%%%%%%%%%%%%%%%%

\section{Additional Ablations and Diagnostics}
\label{app:additional_ablations}

Unless stated otherwise, the following controlled analyses use GR00T-N1.7 on a fixed SimplerEnv-Bridge evaluation run shared across all variants.
The main benchmark tables separately aggregate three evaluation runs.
This section provides the exact control configurations, diagnostic protocols, full quantitative analyses, and additional readouts underlying \Cref{sec:analysis}.
Results already reported in \Cref{tab:supervision-controls-main,fig:intent-structure-main,tab:objective-intervention-main} are referenced directly rather than duplicated.

%%%%%%%%%%%%%%%%%%%%%%%%%%%%%%%%%%%%%%%%
\subsection{Supervision and Capacity Controls}
\label{app:supervision_controls}

We isolate whether the improvement comes from teacher-derived intent, visual and textual grounding supervision, an endpoint-informed latent, or additional latent capacity.
All variants use the same GR00T-N1.7 backbone, Bridge demonstrations, optimization budget, and evaluation protocol.
The groundings-only control retains the visual and textual grounding streams, \(\mathcal L_V\), \(\mathcal L_R\), and grounding self-conditioning while removing the intent queries and intent supervision.
The future-supervision control retains the latent and grounding streams but replaces \(I_t^\star\) with an endpoint visual representation.
The free-latent control retains the latent queries, bottleneck context routing, context dropout, visual and textual grounding streams, alignment objectives, and self-conditioning while removing both \(\mathcal L_I\) and \(\mathcal L_{\mathrm{mis}}\).
The intent-only control retains teacher-intent alignment, bottleneck context routing, context dropout, and action-only intent-mismatch training while removing the visual and textual grounding streams.

\begin{table}[htbp]
\centering
\small
\caption{\textbf{Configuration of supervision and capacity controls.}
Task-level success rates are reported in \Cref{tab:supervision-controls-main}.}
\label{tab:supervision-control-configurations}
\setlength{\tabcolsep}{5pt}
\begin{tabular}{lccc}
\toprule
Variant & Latent target & V/R groundings & Latent queries \\
\midrule
Groundings only & -- & Yes & No \\
Future supervision & Endpoint visual & Yes & Yes \\
Free latent & -- & Yes & Yes \\
Intent only & Teacher intent & No & Yes \\
\method{} & Teacher intent & Yes & Yes \\
\bottomrule
\end{tabular}
\end{table}

As shown in \Cref{tab:supervision-controls-main}, the non-intent controls reach at most \(68.0\%\), whereas intent only reaches \(76.0\%\).
Teacher-derived intent therefore provides a \(14.5\) percentage-point gain over the action-only baseline and an \(8.0\) percentage-point gain over future supervision.
Adding visual and textual groundings further increases average success to \(85.5\%\), showing that they complement rather than replace intent supervision.

\begin{table}[t]
\centering
\caption{\textbf{Supervision and capacity controls on SimplerEnv-Bridge.}
$V$ and $R$ denote the visual and textual grounding streams. All variants share the
backbone, budget, and a fixed evaluation run. Success rates (\%).}
\label{tab:supervision-controls-full}
\small
\setlength{\tabcolsep}{5pt}
\begin{tabular}{@{}lcc|ccccc@{}}
\toprule
Variant & $V$ & $R$ & Spoon & Carrot & Stack & EP-Basket & Avg. \\
\midrule
GR00T-N1.7                 & --        & --        & 82.0 & 68.0 & 66.0 & 30.0  & 61.5 \\
\midrule
\quad + Groundings only    & \ding{51} & \ding{51} & 88.0 & 82.0 & 38.0 & 32.0  & 60.0 \\
\quad + Future supervision & \ding{51} & \ding{51} & 84.0 & 70.0 & 56.0 & 62.0  & 68.0 \\
\quad + Free latent        & \ding{51} & \ding{51} & 78.0 & 76.0 & 48.0 & 26.0  & 57.0 \\
\midrule
\quad + Intent only        & --        & --        & 86.0 & 84.0 & 70.0 & 64.0  & 76.0 \\
\quad + Intent, $V$        & \ding{51} & --        & 90.0   & \textbf{88.0}   & 72.0   & 74.0    & 81.0   \\
\quad + Intent, $R$        & --        & \ding{51} & 90.0   & 86.0   & 70.0   & 72.0    &79.5   \\
\quad \method{} (Ours)     & \ding{51} & \ding{51} & \textbf{90.0} & 78.0 & \textbf{74.0} & \textbf{100.0} & \textbf{85.5} \\
\bottomrule
\end{tabular}
\end{table}

Each grounding stream also contributes on its own, raising average success from \(76.0\%\) to \(81.0\%\) with the visual stream and \(79.5\%\) with the textual stream.
Their joint effect (\(+9.5\)) exceeds the sum of the individual gains (\(+8.5\)), and is concentrated on EP-Basket, where either stream alone reaches roughly \(73\%\) but both together reach \(100.0\%\).
The textual stream helps despite aligning only weakly with its paired target (\Cref{tab:grounding-embedding-readout}), consistent with it acting as a downstream realization rather than a standalone readout.

%%%%%%%%%%%%%%%%%%%%%%%%%%%%%%%%%%%%%%%%
\subsection{Intent-Dependence Mechanism Ablations}
\label{app:mechanism_ablations}

We ablate the two training mechanisms introduced to make downstream decoding depend on recovered intent: bottleneck context routing and intent-mismatch training.
The full model samples \(\alpha\sim\mathrm{Bernoulli}(0.5)\), requiring action and grounding rows to obtain VLA context through the intent rows when \(\alpha=0\), and applies \(\mathcal L_{\mathrm{mis}}\) to encourage lower downstream loss under matched than mismatched intent.
We compare the full model with variants that remove the context bottleneck, the mismatch objective, or both.

\begin{table}[htbp]
\centering
\small
\caption{\textbf{Ablation of intent-dependence mechanisms on SimplerEnv-Bridge.}
All variants use GR00T-N1.7, the same training budget, and the same fixed evaluation run.
The context-bottleneck ablation allows action and grounding rows to access the VLA context throughout training, while the mismatch ablation sets \(\lambda_{\mathrm{mis}}=0\).}
\label{tab:mechanism-ablations}
\setlength{\tabcolsep}{5pt}
\begin{tabular}{lccccc}
\toprule
Variant & Spoon & Carrot & Stack & EP-Basket & Avg. \\
\midrule
\method{} & 90.0 & 78.0 & 74.0 & 100.0 & 85.5 \\
w/o context bottleneck & 88.0 & 76.0 & 60.0 & 56.0 & 70.0 \\
w/o \(\mathcal L_{\mathrm{mis}}\) & 92.0 & 80.0 & 30.0 & 94.0 & 74.0 \\
w/o both & 86.0 & 72.0 & 58.0 & 46.0 & 65.5 \\
\bottomrule
\end{tabular}
\end{table}

Removing the context bottleneck reduces average success from \(85.5\%\) to \(70.0\%\), while removing intent-mismatch training reduces it to \(74.0\%\).
Removing both mechanisms further lowers success to \(65.5\%\).
Thus, both mechanisms contribute substantially and complement one another: bottleneck routing encourages downstream decoding to obtain contextual information through the recovered intent, while \(\mathcal L_{\mathrm{mis}}\) encourages that dependence to be sensitive to the intent's specific content rather than merely to the presence of the latent pathway.

%%%%%%%%%%%%%%%%%%%%%%%%%%%%%%%%%%%%%%%%
\subsection{Functional Role and Structure of Recovered Intent}
\label{app:validating_intent}

We examine whether the recovered representation is used by the action decoder, whether it retains behavior objective and execution progress across different action-sequence realizations, how this information propagates through decoder depth, and whether objective-specific edits affect closed-loop behavior.

\noindent\textbf{The action decoder relies on recovered intent.}
At the intent-alignment layer, we replace all eight recovered intent states with zero vectors while preserving the observation, instruction, proprioceptive state, action noise, grounding states, and remaining decoder computation.

\begin{table}[htbp]
\centering
\small
\caption{\textbf{Dependence on recovered intent.}
The zero-intent intervention replaces all intent states at the alignment layer while preserving all other policy inputs.
Both conditions use the same evaluation run on SimplerEnv-Bridge.}
\label{tab:zero-intent-intervention}
\setlength{\tabcolsep}{6pt}
\begin{tabular}{lccccc}
\toprule
Intent state & Spoon & Carrot & Stack & EP-Basket & Avg. \\
\midrule
\rowcolor{green!15}
\textbf{Recovered intent}
& \textbf{90.0}
& \textbf{78.0}
& \textbf{74.0}
& \textbf{100.0}
& \textbf{85.5} \\
Zero intent
& 58.0
& 44.0
& 40.0
& 40.0
& 45.5 \\
\bottomrule
\end{tabular}
\end{table}

As shown in \Cref{tab:zero-intent-intervention}, zeroing recovered intent reduces average success from \(85.5\%\) to \(45.5\%\), below the \(61.5\%\) action-only baseline.
The recovered state is therefore not merely an auxiliary alignment target but a representation used by downstream action decoding.

\noindent\textbf{Intent represents behavior objectives across domains.}
We collect the eight intent-row hidden states immediately before the alignment layer from 580 real demonstration segments and 202 simulation rollout segments.
Each segment is represented by the flattened \(K_I d_D\)-dimensional recovered intent state.
We reduce the representations to 50 dimensions using PCA and fit three shrinkage-LDA axes from the four Bridge task labels.
The two-dimensional task-discriminative projection is shown in \Cref{fig:intent-objective-main}.
Raw-space one-nearest-neighbor classification achieves \(92.0\%\) task purity, and episode-disjoint shrinkage LDA achieves \(86.2\%\) accuracy.
Randomly permuting the task labels gives \(31.9\%\) mean accuracy across 20 permutations, with a maximum of \(41.4\%\).
A classifier trained only on real-demonstration intents classifies simulation-rollout intents with \(63.9\%\) to \(80.7\%\) accuracy across tasks.
These results show that objective information persists across episodes and transfers from real demonstrations to simulation rollouts.

\noindent\textbf{Intent captures progress shared across related manipulation tasks.}
The four Bridge tasks manipulate different objects and target relations but share an approach--grasp--transport--place structure.
We divide each episode into early, middle, and late intervals and fit a phase-discriminative projection using labels pooled across tasks.
As shown in \Cref{fig:intent-progress-main}, segments from different tasks organize by execution stage.
Three-way phase prediction achieves \(77.8\%\) five-fold cross-validation accuracy, compared with a \(33.3\%\) chance level.
Task purity within the phase plane is \(36.6\%\), while task identity remains readable within the early, middle, and late intervals at \(94.0\%\), \(90.0\%\), and \(90.0\%\), respectively.
We also fit a continuous progress direction with ridge regression and evaluate it in a leave-one-task-out setting.
The held-out correlations are 0.62 for Spoon, 0.70 for Carrot, 0.72 for Stack, and 0.69 for EP-Basket, compared with an in-sample correlation of 0.88.

\begin{table}[htbp]
\centering
\footnotesize
\caption{\textbf{Objective and task-family-shared progress structure in recovered intent.} The supervision control applies the same probe to the free-latent variant, which receives identical context access but no teacher-intent supervision.}
\label{tab:intent-factorization}
\setlength{\tabcolsep}{5pt}
\begin{tabularx}{0.94\linewidth}{@{}>{\raggedright\arraybackslash}X>{\raggedright\arraybackslash}Xc@{}}
\toprule
Property & Evaluation & Result \\
\midrule
\multicolumn{3}{l}{\textit{Behavior objective}} \\
Objective identity & Raw-space task 1-NN purity & 92.0\% \\
Objective generalization & Episode-disjoint task LDA & 86.2\% \\
Supervision control & Same probe, free latent (no \(\mathcal{L}_I\)) & 55.7\% \\
Cross-domain transfer & Train on real, test on simulation & 63.9--80.7\% \\
Objective retention & Task purity in early/middle/late phases & 94.0/90.0/90.0\% \\
\midrule
\multicolumn{3}{l}{\textit{Shared execution progress}} \\
Phase readability & Three-way phase prediction & 77.8\% \\
Task leakage & Task purity within phase plane & 36.6\% \\
Cross-task transfer & Leave-one-task-out progress correlation & 0.62--0.72 \\
\bottomrule
\end{tabularx}
\end{table}

Together, \Cref{fig:intent-structure-main,tab:intent-factorization} show complementary objective and progress structure in the recovered representation.
The task-discriminative structure captures which object-level objective is being pursued, while the phase-discriminative structure captures progress shared by this family of manipulation tasks.
We interpret the latter as a task-family-shared progress coordinate rather than a universal progress axis for arbitrary robot skills.

\noindent\textbf{Intent captures objective beyond action-sequence realization.}
\label{app:intent_action_sequence}
We distinguish intent from skill by operationalizing skill as the executed action sequence.
The action sequence describes how a behavior is realized, whereas intent denotes the local objective that the behavior serves.
To test whether intent merely summarizes the executed sequence, we compare progress-matched episode pairs under two opposing conditions: the same objective realized by dissimilar sequences, and different objectives realized by similar sequences.
We use episode-disjoint discovery and test splits, with all centering statistics and pair-selection thresholds estimated from discovery episodes.

\begin{figure*}[htbp]
\centering
\includegraphics[width=\textwidth]{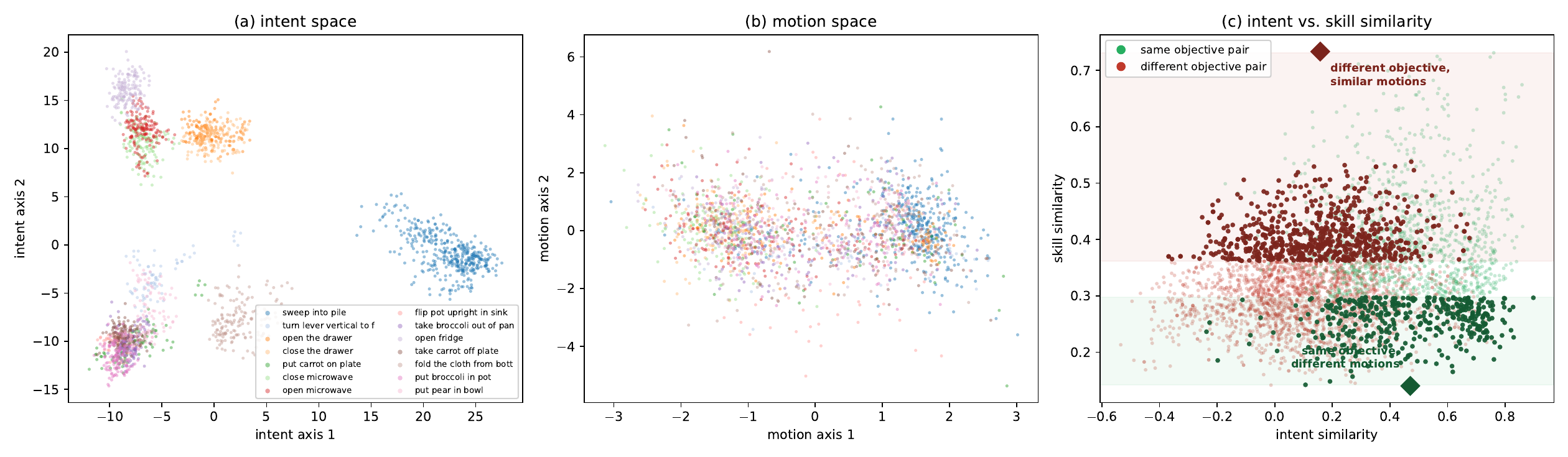}
\caption{\textbf{Intent captures objective beyond action-sequence realization.}
We operationalize skill as the executed action sequence.
(a) Teacher intent targets separate 14 behavior objectives on held-out episodes.
(b) A matched supervised projection of local action snippets yields weak objective separation.
(c) In a projection-free comparison of progress-matched episode pairs, the same objective realized by dissimilar action sequences remains more similar in intent space than different objectives realized by similar sequences.
Light points show all eligible pairs, dark points show the selected contrast sets, and diamonds mark their means.
Recovered policy intent exhibits the same ordering.}
\label{fig:intent-vs-action-sequence}
\end{figure*}

As shown in \Cref{fig:intent-vs-action-sequence}, teacher intent targets separate by objective on held-out episodes, while a matched supervised projection of local action snippets yields weak objective separation.
More importantly, the projection-free comparison shows that recovered-intent similarity remains higher for the same objective with dissimilar sequences than for different objectives with similar sequences (\(0.150\) vs.\ \(-0.017\)), with a gap of \(0.167\) and a 95\% episode-bootstrap confidence interval of \([0.089,0.245]\).
Teacher targets show the same ordering (\(0.470\) vs.\ \(0.158\); gap \(0.312\), 95\% CI \([0.253,0.375]\)).
Thus, the representation preserves what the behavior is meant to accomplish across changes in how it is executed, distinguishing intent from a latent code that merely summarizes the action sequence.

\noindent\textbf{Task information propagates from intent into the grounding streams.}
A one-time edit at the intent-alignment layer may be overwritten if later blocks reconstruct the original objective from other token streams.
We therefore measure one-nearest-neighbor task purity for the state, action, visual-grounding, textual-grounding, and intent rows throughout the 32 decoder blocks.

\begin{table}[htbp]
\centering
\small
\caption{\textbf{Task information across decoder rows and depth.}
The table summarizes task information before and after the intent-alignment layer.}
\label{tab:row-probe}
\setlength{\tabcolsep}{6pt}
\begin{tabular}{lcc}
\toprule
Decoder rows & Blocks \(0\)--\(15\) & Blocks \(19\)--\(31\) \\
\midrule
Intent & \(0.63 \rightarrow \mathbf{0.93}\) & \(0.77 \pm 0.05\) \\
Visual grounding & Near chance & Up to \(\mathbf{0.95}\) \\
Textual grounding & Near chance & Up to \(\mathbf{0.97}\) \\
Action and state & Near chance & Near chance \\
\bottomrule
\end{tabular}
\end{table}

Intent-row task purity peaks at 0.93 immediately before the alignment layer.
After alignment, task information appears in the visual and textual grounding rows, reaching 0.95 and 0.97, while the action and state rows remain near chance.
This pattern motivates editing both intent and grounding rows throughout the decoder tail.

\noindent\textbf{Editing objective-specific coordinates changes closed-loop behavior.}
Zeroing intent establishes that the pathway is necessary but does not isolate the effect of its objective content.
We therefore edit only the task-discriminative coordinates while preserving the remaining representation.
Let \(g\in\{I,G\}\) index the edited token group, where \(I\) contains the intent rows and \(G=V\oplus R\) contains the visual and textual grounding rows.
Let \(h_{g,k}\in\mathbb R^{p_g}\) denote the flattened representation after decoder block \(k\).
For task \(t\), we compute the block-wise centroid
\[
\mu_{g,k}^{(t)}
=
\mathbb E\!\left[h_{g,k}\mid \operatorname{task}=t\right]
\]
and form a centered matrix from the four task centroids.
The columns of \(U_{g,k}\) are the right singular vectors spanning the resulting between-task subspace.
For an injected task \(t_d\), we replace only the coordinates inside this subspace:
\[
h_{g,k}^{\prime}
=
h_{g,k}
+
U_{g,k}
\left[
U_{g,k}^{\top}\mu_{g,k}^{(t_d)}
-
U_{g,k}^{\top}h_{g,k}
\right].
\]
This sets the task-discriminative coordinates to those of the injected objective while preserving the orthogonal component.
We apply the edit to the intent and grounding groups after every decoder block from 16 through 24, without modifying the action or state rows.
Task centroids and bases are estimated from separate rollouts under environment seeds disjoint from intervention evaluation.
Each same-objective, cross-objective, and Gaussian-noise condition uses the same intervention schedule and is evaluated over 50 episodes.
Donors are captured from separate rollouts, so the same-objective condition also receives an injection of equal strength and differs from the cross-objective condition only in donor content.

The resulting closed-loop success matrix is reported in \Cref{tab:objective-intervention-main}.
Same-objective interventions retain \(84.5\%\) success on average, compared with \(45.2\%\) for cross-objective interventions, while Gaussian corruption reduces success to \(1.0\%\).
The dependence on which objective is injected, together with the contrast against unstructured noise, shows that the intent-mediated pathway carries content rather than acting as an undifferentiated gate.

\begin{figure}[htbp]
\centering
\includegraphics[width=0.4\linewidth]{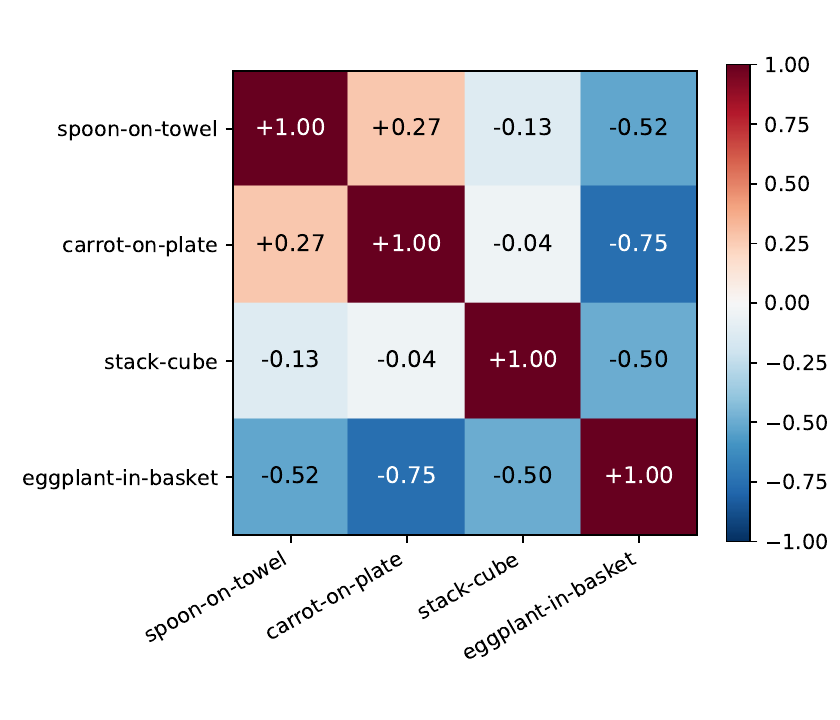}
\caption{\textbf{Task relatedness in recovered-intent space.}
Each entry reports centered cosine similarity between the recovered-intent centroids of two SimplerEnv tasks.
Spoon and Carrot form the closest pair, while EP-Basket is the most separated from the other tasks.}
\label{fig:task-intent-similarity}
\end{figure}

\noindent\textbf{Intent geometry reflects task relatedness.}
\label{app:task_intent_similarity}
The uneven effects of cross-objective intervention raise the question of whether the recovered intent space reflects which tasks are related.
We therefore compare centered cosine similarity between recovered-intent centroids for the four SimplerEnv evaluation tasks.
As shown in \Cref{fig:task-intent-similarity}, Spoon and Carrot form the closest pair, Stack is more distinct, and EP-Basket is the most separated from the other tasks.
This organization is consistent with Spoon and Carrot sharing a closely related tabletop manipulation structure, whereas EP-Basket differs in both its objective and scene.
The result provides a descriptive representation-level notion of task relatedness that may help explain why some cross-objective transfers are more compatible than others.
Because scene and objective differences are not independently controlled, we do not interpret the similarity as a causal measure of transfer.

\noindent\textbf{Editing stage-specific coordinates changes which stage is executed.}
We repeat the procedure with \(U_{g,k}\) spanning the between-phase subspace, estimated from early, middle, and late centroids pooled across tasks using the projection of \Cref{fig:intent-progress-main}.
All other aspects of the protocol are unchanged, so the two interventions differ only in which subspace is replaced.
Forcing an early-stage donor reduces average success to \(7.3\%\), and forcing a late-stage donor reduces it to \(5.5\%\), both approaching the \(1.0\%\) obtained under Gaussian corruption.
The phase-matched control, which injects a donor from the same task at the receiver's own stage, retains \(50\%\), \(70\%\), and \(32\%\) on Spoon, Carrot, and Stack.
The gap between this control and the uninjected policy reflects the cost of clamping the intent stream over the window, which varies by task.

\begin{figure*}[htbp]
\centering
\includegraphics[width=\textwidth]{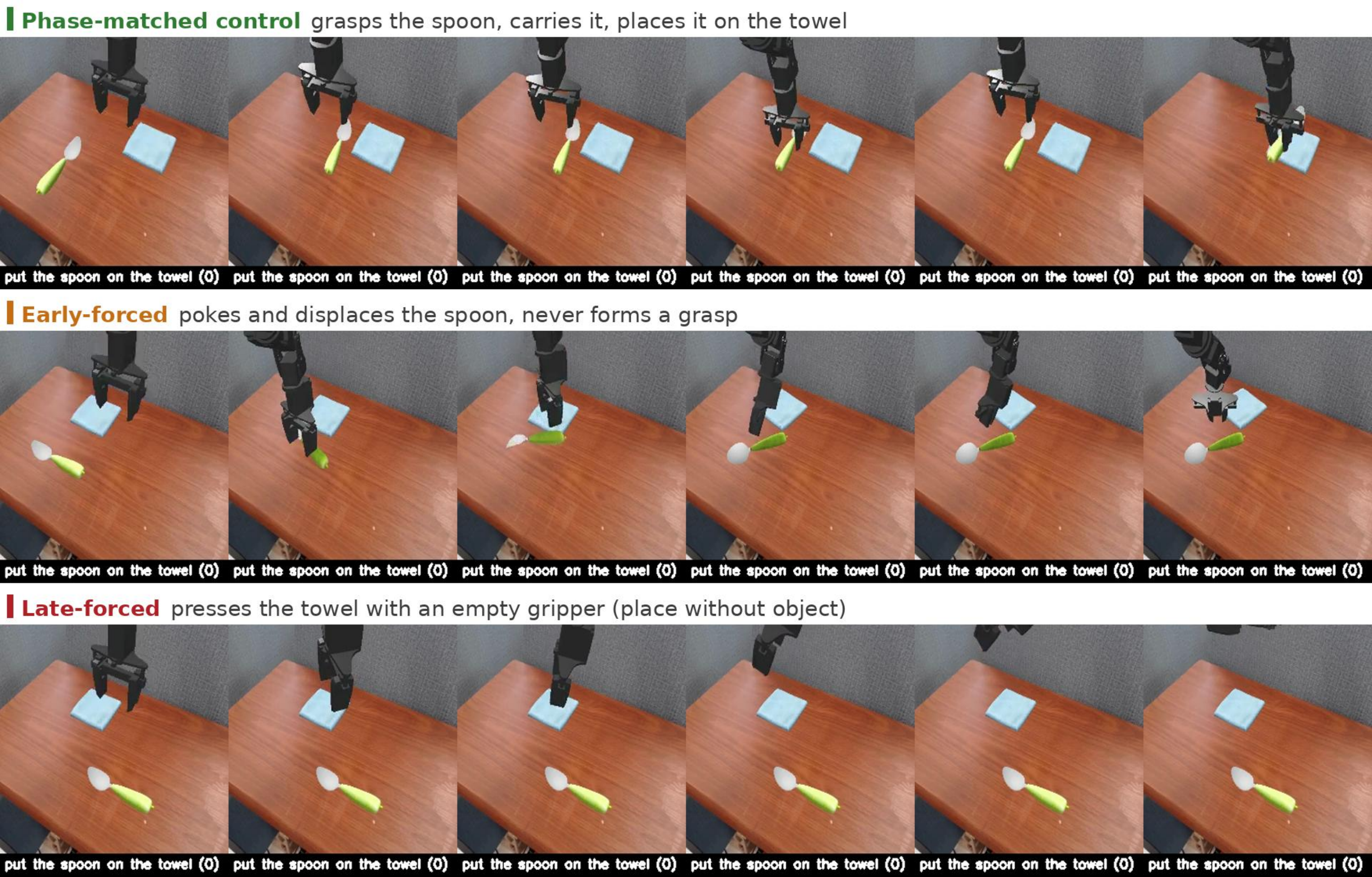}
\caption{\textbf{Stage-forced intervention on Spoon.}
Rows show phase-matched, early-forced, and late-forced injection.
Early forcing leaves the policy displacing the spoon without ever closing the gripper.
Late forcing skips the grasp entirely: the policy moves to the towel and presses down with an empty gripper, executing placement as if it were already carrying the object.}
\label{fig:stage-forced-spoon}
\end{figure*}

\begin{figure*}[htbp]
\centering
\includegraphics[width=\textwidth]{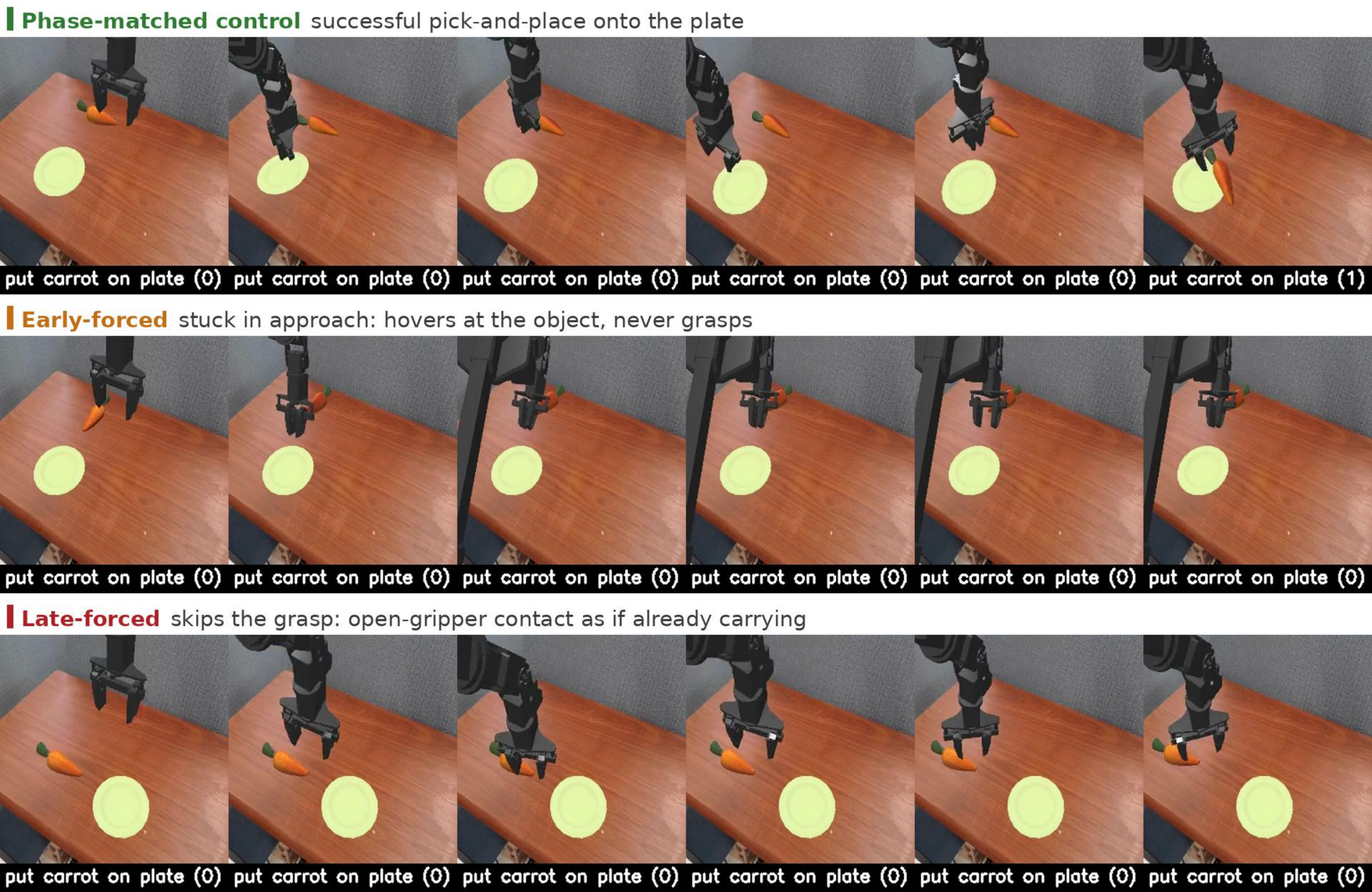}
\caption{\textbf{Stage-forced intervention on Carrot.}
Early forcing leaves the policy hovering over the carrot without transitioning to grasp.
Late forcing produces open-gripper contact, as if the object were already being transported.}
\label{fig:stage-forced-carrot}
\end{figure*}

\begin{figure*}[htbp]
\centering
\includegraphics[width=\textwidth]{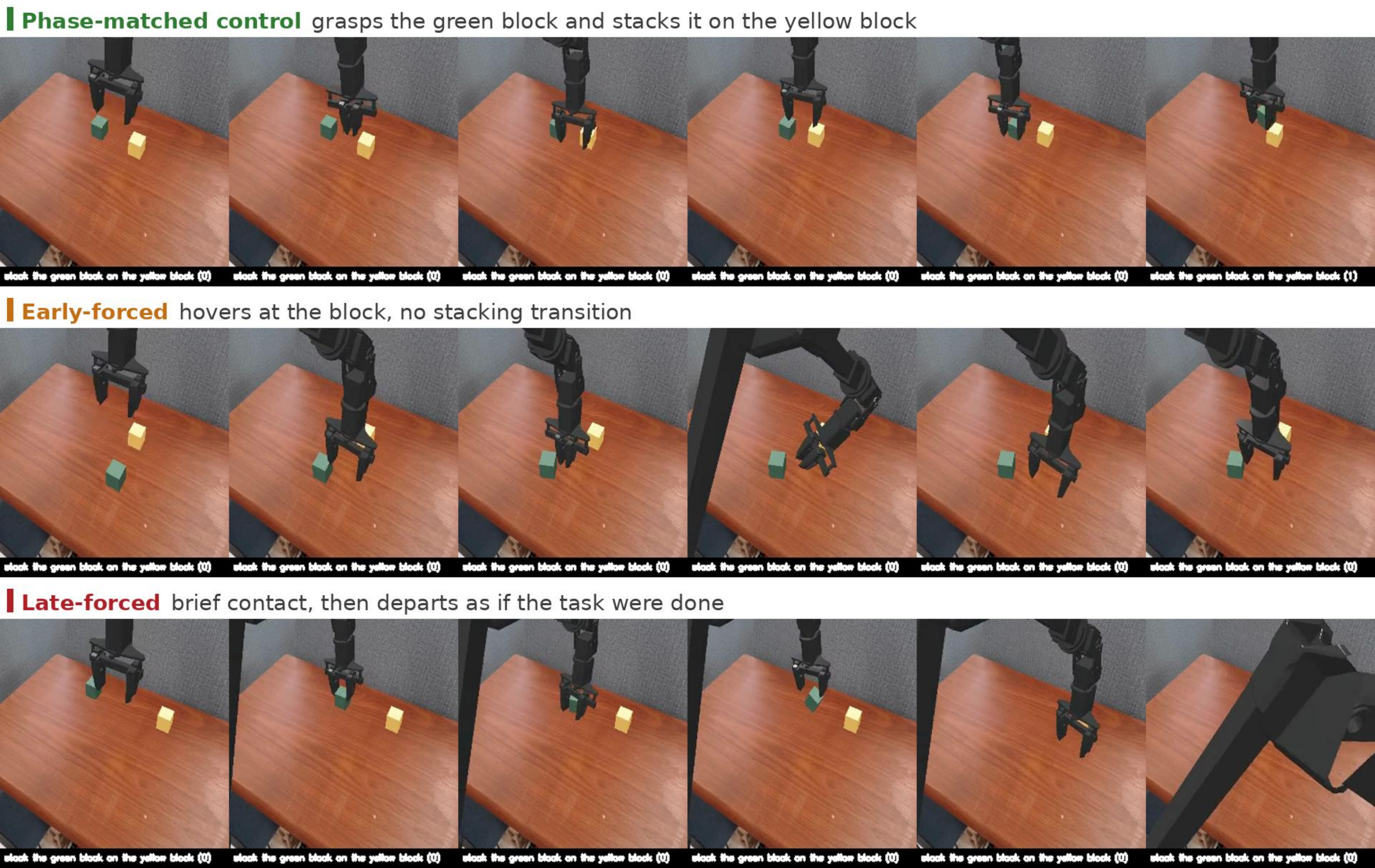}
\caption{\textbf{Stage-forced intervention on Stack.}
Early forcing leaves the policy fumbling at the green block without initiating the stack.
Late forcing produces brief contact followed by departure, as if the task were already complete.}
\label{fig:stage-forced-stack}
\end{figure*}

As shown in \Cref{fig:stage-forced-spoon,fig:stage-forced-carrot,fig:stage-forced-stack}, rollouts under stage-forced injection do not fail incoherently but execute behavior consistent with the injected stage.
Early forcing leaves the policy executing approach behavior without transitioning to grasp: it hovers over the target on Carrot, displaces the spoon without closing the gripper on Spoon, and fumbles at the green block without initiating the stack on Stack.
Late forcing produces the opposite pattern, skipping the grasp and executing placement.
This is clearest on Spoon, where the policy moves to the towel and presses down with an empty gripper.
Since both conditions apply injections of equal strength and differ only in donor content, the contrast is attributable to the stage information carried by the intent pathway rather than to perturbation magnitude.

%%%%%%%%%%%%%%%%%%%%%%%%%%%%%%%%%%%%%%%%
\subsection{Visual and Textual Readout of Intent}
\label{app:intent_readout}

\Cref{app:supervision_controls} identifies which supervision improves policy performance, whereas this subsection localizes the resulting behavior information within the decoder.
We freeze the trained policy for all readout experiments.
Let
\[
\widehat{I}_t
=
P_I\!\left(H_{I,t}^{L_{\mathrm{tap}}}\right),
\qquad
\widehat{V}_t
=
P_V\!\left(H_{V,t}^{L_{\mathrm{out}}}\right),
\qquad
\widehat{R}_t
=
P_R\!\left(H_{R,t}^{L_{\mathrm{out}}}\right)
\]
denote the projected intent and the final visual and textual grounding predictions.
We evaluate them using nearest-neighbor retrieval and separately trained readout decoders while keeping the policy frozen.

\noindent\textbf{Embedding-space readout.}
We extract \(\widehat{V}_t\), \(\widehat{R}_t\), \(V_t^\star\), and \(R_t^\star\) for approximately 10,000 held-out segments.
We subtract the dataset mean before computing cosine similarity and exclude retrieval candidates from the same episode.

\begin{table*}[htbp]
\centering
\small
\caption{\textbf{Embedding-space readout of the visual and textual groundings.}
Correct cosine measures centered similarity to the paired target.
Shuffled cosine uses randomly permuted targets.
Median rank is zero-indexed.}
\label{tab:grounding-embedding-readout}
\setlength{\tabcolsep}{6pt}
\begin{tabular}{lcccccc}
\toprule
Readout
& Correct cosine
& Shuffled cosine
& Margin
& Top-1
& Top-10
& Median rank \\
\midrule
\(\widehat{V}_t \rightarrow V_t^\star\)
& \textbf{0.650}
& 0.003
& \textbf{0.647}
& \textbf{72.6\%}
& \textbf{95.3\%}
& \textbf{0} \\
\(\widehat{R}_t \rightarrow R_t^\star\)
& 0.158
& \(-0.004\)
& 0.161
& 1.5\%
& 5.0\%
& 1211 \\
\bottomrule
\end{tabular}
\end{table*}

The visual grounding provides a highly readable endpoint representation, retrieving the paired target at rank one in \(72.6\%\) of held-out segments and within the top ten in \(95.3\%\).
The textual grounding is more similar to its paired target than to shuffled targets, although its standalone nearest-neighbor geometry is less concentrated.
Purpose semantics are most compactly represented in \(\widehat I_t\), while the textual grounding rows provide a downstream semantic realization used during joint decoding.

\noindent\textbf{Visual grounding depends on intent.}
We repeat the visual readout after zeroing recovered intent or replacing it with recovered intent from another task while preserving the remaining policy inputs.

\begin{table}[htbp]
\centering
\small
\caption{\textbf{Effect of intent intervention on visual grounding.}}
\label{tab:visual-grounding-intervention}
\setlength{\tabcolsep}{8pt}
\begin{tabular}{lcc}
\toprule
Intent condition & Centered margin & Top-1 retrieval \\
\midrule
Recovered intent & \textbf{0.647} & \textbf{72.6\%} \\
Zero intent & 0.513 & 59.1\% \\
Cross-task intent & 0.393 & 41.8\% \\
\bottomrule
\end{tabular}
\end{table}

Visual readout quality decreases from recovered to zero to cross-task intent.
The predicted visual outcome therefore depends on the content supplied through the intent pathway.

\noindent\textbf{Visual-outcome retrieval.}
We retrieve actual endpoint frames using \(\widehat V_t\).
For the intervention montage, the retrieval bank includes the designated current-task and injected-task endpoints.

\begin{figure*}[htbp]
\centering
\includegraphics[width=0.95\textwidth]{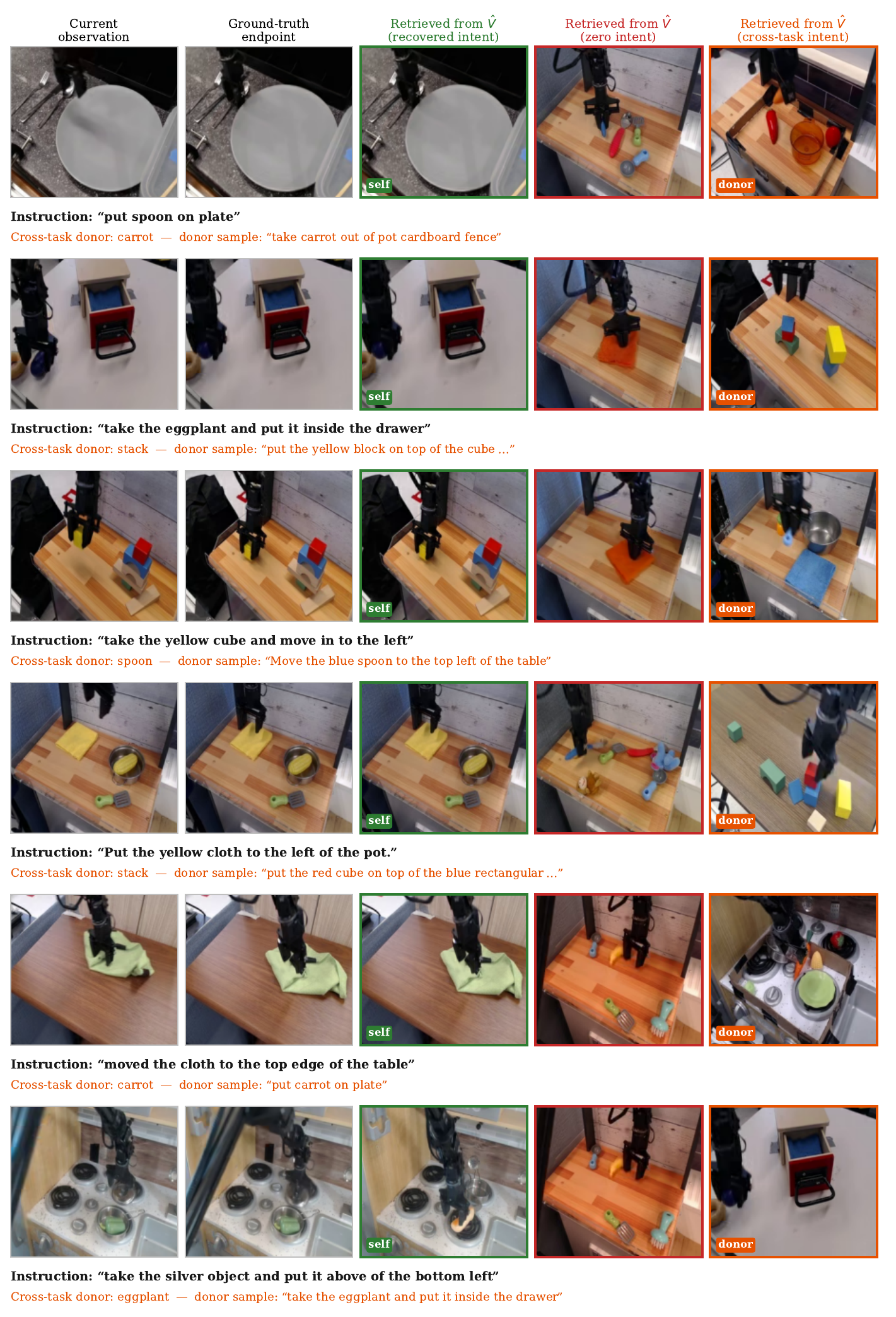}
\caption{\textbf{Non-parametric visual-outcome readout.}
Each row shows the current observation, ground-truth endpoint, and endpoint retrieved from \(\widehat V_t\) under recovered, zero, and cross-task intent.
Recovered intent retrieves the current behavior outcome, while cross-task intent redirects retrieval toward the injected behavior.
Across the evaluated pairs, recovered intent retrieves the current endpoint in 70.5\% of cases.
After cross-task replacement, the injected endpoint is retrieved in 78.6\% of cases, while the original endpoint remains in only 0.1\%.}
\label{fig:visual-outcome-retrieval}
\end{figure*}

The examples in \Cref{fig:visual-outcome-retrieval} show that recovered intent preserves the current object, scene, and resulting relation, while cross-task intent redirects the predicted outcome toward the injected behavior.

\noindent\textbf{Learned visual and textual decoding.}
We additionally train lightweight decoders to test whether the representations can be converted directly into an endpoint image and a functional-purpose statement.
The visual decoder maps the 16 ordered visual slots to the endpoint RGB image using a transformer bottleneck and convolutional upsampling.
The policy is frozen, and the decoder is trained on visual targets \(V_t^\star\) and endpoint observations.
On held-out data, reconstruction from \(V_t^\star\) obtains a changed-region \(L_1\) error of 0.0819, compared with 0.3042 for copying the current observation.
For textual readout, we compare the final textual grounding \(\widehat R_t\) with the projected intent \(\widehat I_t\).
The intent decoder is trained on approximately 9,900 unmodified intent--statement pairs using an episode-disjoint split.
We report deterministic object F1, state-change verb F1, and ROUGE-L.

\begin{table}[htbp]
\centering
\small
\caption{\textbf{Textual-purpose readout.}
The oracle row decodes the teacher textual target.
The remaining rows decode the predicted textual grounding and recovered intent.}
\label{tab:textual-purpose-readout}
\setlength{\tabcolsep}{7pt}
\begin{tabular}{lccc}
\toprule
Readout representation & Object F1 & Verb F1 & ROUGE-L \\
\midrule
\(R_t^\star\) (oracle target) & \textbf{0.998} & \textbf{1.000} & \textbf{0.933} \\
\(\widehat{R}_t\) & 0.314 & 0.195 & 0.255 \\
\(\widehat{I}_t\) & \textbf{0.693} & \textbf{0.432} & \textbf{0.380} \\
\bottomrule
\end{tabular}
\end{table}

The recovered intent is substantially more readable than the final textual grounding, improving object F1 from 0.314 to 0.693 and verb F1 from 0.195 to 0.432.
This localizes the semantic bottleneck primarily in recovered intent while retaining the textual grounding as a downstream representation used during joint decoding.

\begin{figure*}[htbp]
\centering
\includegraphics[width=\textwidth]{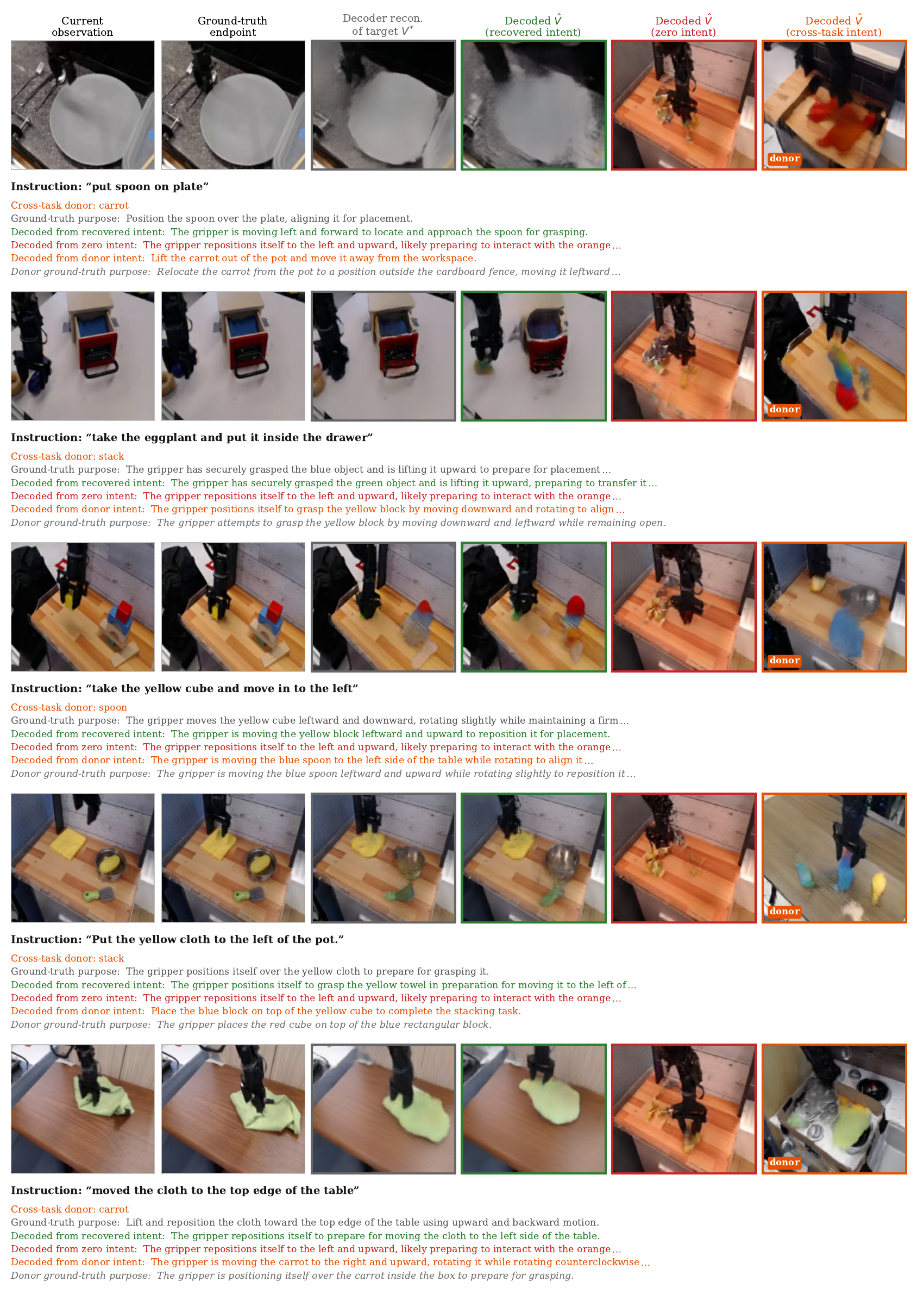}
\caption{\textbf{Learned visual and textual readout of intent.}
Each row shows the current observation, ground-truth endpoint, visual reconstruction from \(V_t^\star\), and visual outcomes decoded from \(\widehat V_t\) under recovered, zero, and cross-task intent.
The text below each row shows the ground-truth purpose and statements decoded from recovered, zero, and injected intent.
Recovered intent produces an outcome and purpose consistent with the current behavior, while cross-task intent redirects both modalities toward the injected behavior.}
\label{fig:joint-intent-decoding}
\end{figure*}

The visual and textual readouts change consistently under intent intervention.
These results support the intended hierarchy of \method{}, in which intent stores a compact behavior-level representation and visual outcome and textual purpose provide complementary observable realizations.

% \subsection{Information-Flow Ablation}
% \label{app:design_choices}

% We analyze the asymmetric information-flow pattern used to organize action and grounding prediction around recovered intent.
% The asymmetric design allows action rows to read the recovered intent and both grounding streams while preventing grounding rows from directly reading the noised action stream.
% We compare it with a symmetric suffix-attention variant in which grounding and action rows can exchange information.
% Both variants use \(\pi_{0.5}\) under the same training and evaluation protocol.

% \begin{table}[htbp]
% \centering
% \small
% \caption{\textbf{Information-flow ablation on SimplerEnv-Bridge.}
% Both configurations use \(\pi_{0.5}+\method{}\) and the same evaluation run.}
% \label{tab:information-flow}
% \setlength{\tabcolsep}{7pt}
% \begin{tabular}{lcc}
% \toprule
% Configuration & Success (\%) & Improvement \\
% \midrule
% Symmetric grounding attention & 41.8 & -- \\
% \rowcolor{green!15}
% Asymmetric information flow & \textbf{49.0} & \(+7.2\) \\
% \bottomrule
% \end{tabular}
% \end{table}

% As shown in \Cref{tab:information-flow}, asymmetric information flow improves success from \(41.8\%\) to \(49.0\%\), supporting the design choice of separating grounding prediction from the noised action stream.

%%%%%%%%%%%%%%%%%%%%%%%%%%%%%%%%%%%%%%%%
\subsection{Teacher Quality and Intent-Target Geometry}
\label{app:teacher_target_consistency}

We examine whether intent distillation benefits from an arbitrary teacher, or whether the teacher target itself must have particular structure.
We repeat the controlled SimplerEnv-Bridge training while changing only the teacher cache, keeping the demonstrations, student architecture, optimization budget, checkpoint selection, and evaluation protocol fixed.

\noindent\textbf{Teacher choice matters.}
Table~\ref{tab:teacher-policy-transfer} shows that teacher choice substantially changes downstream performance.
Qwen3.5-9B~\citep{qwen2026qwen35} reduces average success to \(54.5\%\), below the \(61.5\%\) action-only baseline, whereas Qwen3-VL-8B-Instruct~\citep{bai2025qwen3vl} reaches \(73.5\%\) and Cosmos-Reason2-8B reaches \(85.5\%\).
Since Cosmos-Reason2-8B~\citep{nvidia2025cosmosreason2code} is obtained by physical-AI post-training of Qwen3-VL-8B-Instruct, this pair provides a controlled comparison of the effect of teacher-side post-training.

\begin{table}[htbp]
\centering
\small
\caption{\textbf{Controlled teacher swap on SimplerEnv-Bridge.}
All variants use the same GR00T-N1.7 student, demonstrations, optimization budget, and fixed evaluation run; only the teacher cache is changed.}
\label{tab:teacher-policy-transfer}
\setlength{\tabcolsep}{4.5pt}
\begin{tabular}{lccccc}
\toprule
Teacher & Spoon & Carrot & Stack & EP-Basket & Avg. \\
\midrule
No intent distillation & 82.0 & 68.0 & 66.0 & 30.0 & 61.5 \\
Qwen3.5-9B & 84.0 & 78.0 & 40.0 & 16.0 & 54.5 \\
Qwen3-VL-8B-Instruct & 90.0 & 98.0 & 34.0 & 72.0 & 73.5 \\
Cosmos-Reason2-8B & 90.0 & 78.0 & 74.0 & 100.0 & \textbf{85.5} \\
\bottomrule
\end{tabular}
\end{table}

\noindent\textbf{The difference is visible in the intent targets.}
We characterize each teacher's \(I^\star\) targets on the same \(17{,}288\) valid Bridge samples.
The \emph{angular budget} measures the mean cosine distance of L2-normalized targets from their mean direction, and therefore how much sample-dependent directional variation is available to the cosine-alignment objective.
The \emph{effective rank} summarizes the spectrum of the centered target distribution, while the \emph{within-episode share} measures the fraction of centered variance that occurs within episodes rather than between them.

\begin{table}[htbp]
\centering
\small
\caption{\textbf{Intent-target geometry across teachers.}
All statistics use the same valid Bridge samples (\(n=17{,}288\)).
Bridge Avg. reports the corresponding controlled policy performance.}
\label{tab:teacher-target-geometry}
\setlength{\tabcolsep}{4.5pt}
\begin{tabular}{lcccc}
\toprule
Teacher & Angular budget & Eff. rank & Within-ep. share & Bridge Avg. \\
\midrule
Qwen3.5-9B & 0.0027 & 139.4 & 0.308 & 54.5 \\
Qwen3-VL-8B-Instruct & 0.0098 & 7.2 & 0.693 & 73.5 \\
Cosmos-Reason2-8B & 0.0327 & 3.4 & 0.723 & \textbf{85.5} \\
\bottomrule
\end{tabular}
\end{table}

Qwen3.5 produces an almost degenerate alignment signal: its angular budget is only \(0.0027\), so a nearly constant prediction can satisfy much of the cosine-alignment objective, and only \(30.8\%\) of its residual variation occurs within episodes.
Qwen3-VL instead exhibits substantially more sample-dependent and state-resolved variation and recovers a \(12.0\)-percentage-point gain over the action-only baseline.
Cosmos preserves the high within-episode share while increasing the angular budget by \(3.3\times\) over Qwen3-VL, accompanied by a further \(12.0\)-point downstream gain.
The high effective rank of Qwen3.5 therefore does not indicate a useful teacher by itself; it describes the structure of a very small residual signal.

Across the three teachers, angular budget and within-episode share increase monotonically with downstream success.
Given the small number of teachers, we treat this as a diagnostic condition rather than a predictive law: useful intent supervision requires a non-degenerate target whose variation resolves moment-to-moment behavior state rather than primarily static episode identity.
The causal interpretation is limited to the Qwen3-VL--Cosmos pair, where physical-AI post-training is the controlled teacher-side change.

%%%%%%%%%%%%%%%%%%%%%%%%%%%%%%%%%%%%%%%%
\subsection{Intent-Target Source Ablation}
\label{app:intent_target_source}

The default intent target is extracted from the intermediate hidden states over the multimodal teacher-input span \(\mathcal E_t\).
We compare this evidence-side target with the generated-response target \(I_t^{\star,\mathrm{resp}}\), which pools the intermediate hidden states over the complete teacher response \(\mathcal Y_t\), including its reasoning and final functional-purpose statement.
Both variants use the same teacher, \(K_I=8\) target slots, student architecture, grounding targets, optimization budget, and evaluation protocol.
The comparison therefore isolates whether policy supervision is more effective when intent is represented during multimodal evidence processing or after the teacher has synthesized that evidence into a generated response.

\begin{table}[htbp]
\centering
\small
\caption{\textbf{Intent-target source ablation on SimplerEnv-Bridge.}
Evidence-side and generated-response targets use the same teacher layer and \(K_I=8\) contiguous-region pooling.}
\label{tab:intent-target-source}
\setlength{\tabcolsep}{5pt}
\begin{tabular}{lccccc}
\toprule
Intent target & Spoon & Carrot & Stack & EP-Basket & Avg. \\
\midrule
Evidence-side input span \(\mathcal E_t\) & 90.0 & 78.0 & 74.0 & 100.0 & 85.5 \\
Generated-response span \(\mathcal Y_t\) & 76.0 & 86.0 & 76.0 & 80.0 & 79.5 \\
\bottomrule
\end{tabular}
\end{table}

The generated-response target obtains \(79.5\%\) average success, compared with \(85.5\%\) for the evidence-side multimodal target.
Although the generated-response target substantially outperforms the \(61.5\%\) action-only baseline, the evidence-side target provides a further \(6.0\) percentage-point gain.
This result indicates that both target sources provide useful behavior-level supervision, while directly distilling the representation formed during multimodal evidence processing is more effective than using the representation formed after reasoning and response generation.

%%%%%%%%%%%%%%%%%%%%%%%%%%%%%%%%%%%%%%%%%%%%%%%%%%%%%%%%%%%%%%%%%%%%%%%%%%%%%%%%
%%%%%%%%%%%%%%%%%%%%%%%%%%%%%%%%%%%%%%%%%%%%%%%%%%%%%%%%%%%%%%%%%%%%%%%%%%%%%%%%

\section{Additional Results}
\label{app:additional_results}

%%%%%%%%%%%%%%%%%%%%%%%%%%%%%%%%%%%%%%%%
\subsection{Full Quantitative Results}
\label{app:full_results}

We provide complete RoboCasa Kitchen results to further examine the generality of \method{} across VLA backbones and manipulation tasks.
\Cref{tab:robocasa-full} reports the per-task success rates summarized by task category in the main text.
The table compares \method{} with controlled GR00T-N1.7 and \(\pi_{0.5}\) baselines, the
GR00T-N1.7 future-supervision variant, and off-the-shelf GR00T checkpoints trained with larger demonstration sets.

The future-supervision variant exhibits substantially larger per-task variation than \method{}, improving markedly on some tasks (e.g., Open Double Door, Turn Off Stove) while degrading on others where the remaining variants are near ceiling (e.g., Turn On Sink Faucet,
Coffee Press Button). 
Its category and overall averages therefore reflect a redistribution of per-task performance rather than a uniform improvement.

\begin{table*}[htbp]
\centering
\scriptsize
\caption{\textbf{Full RoboCasa Kitchen per-task results.}
Success rates are reported across all 24 tasks.
Controlled GR00T-N1.7 variants and \(\pi_{0.5}\) results report mean \(\pm\) sample standard deviation over three evaluation runs.
The GR00T \(G_{3000}\) columns provide single-value data-scale references.
Bold marks the best controlled result within each backbone in the category-summary rows.}
\label{tab:robocasa-full}
\setlength{\tabcolsep}{3pt}
\resizebox{\textwidth}{!}{%
\begin{tabular}{lccccccc}
\toprule
& \multicolumn{5}{c}{GR00T} & \multicolumn{2}{c}{\(\pi_{0.5}\)} \\
\cmidrule(lr){2-6}
\cmidrule(lr){7-8}
Task
& N1.6 \(G_{3000}\)
& N1.7 \(G_{3000}\)
& N1.7 \(G_{100}\)
& \shortstack{N1.7 + future\\supervision \(G_{100}\)}
& \shortstack{N1.7 + \method{}\\\(G_{100}\)}
& Baseline
& + \method{} \\
\midrule
\multicolumn{8}{l}{\textit{Pick-and-Place}} \\
PnP from Cab to Counter
& 41.0
& 65.0
& \meanstd{26.0}{16.0}
& \meanstd{44.7}{1.2}
& \meanstd{50.0}{4.0}
& \meanstd{18.7}{2.3}
& \meanstd{10.7}{10.1} \\
PnP from Counter to Cab
& 47.5
& 60.0
& \meanstd{37.3}{3.1}
& \meanstd{35.3}{5.0}
& \meanstd{58.7}{10.3}
& \meanstd{18.7}{2.3}
& \meanstd{23.3}{7.6} \\
PnP from Counter to Microwave
& 19.0
& 30.0
& \meanstd{22.0}{5.3}
& \meanstd{57.3}{5.8}
& \meanstd{28.0}{0.0}
& \meanstd{6.7}{2.3}
& \meanstd{6.7}{2.3} \\
PnP from Counter to Sink
& 46.0
& 60.0
& \meanstd{54.7}{11.7}
& \meanstd{29.3}{11.4}
& \meanstd{53.3}{10.3}
& \meanstd{10.0}{7.2}
& \meanstd{16.0}{4.0} \\
PnP from Counter to Stove
& 63.2
& 60.0
& \meanstd{46.7}{7.0}
& \meanstd{56.0}{6.0}
& \meanstd{50.7}{9.9}
& \meanstd{12.0}{6.9}
& \meanstd{8.0}{0.0} \\
PnP from Microwave to Counter
& 24.5
& 19.0
& \meanstd{23.3}{3.1}
& \meanstd{26.7}{2.3}
& \meanstd{32.0}{2.0}
& \meanstd{8.0}{0.0}
& \meanstd{8.0}{0.0} \\
PnP from Sink to Counter
& 50.0
& 65.0
& \meanstd{50.7}{8.3}
& \meanstd{64.7}{9.9}
& \meanstd{62.7}{3.1}
& \meanstd{24.7}{9.9}
& \meanstd{21.3}{2.3} \\
PnP from Stove to Counter
& 54.5
& 65.0
& \meanstd{54.7}{4.2}
& \meanstd{62.0}{3.5}
& \meanstd{62.7}{8.1}
& \meanstd{13.3}{2.3}
& \meanstd{28.7}{4.2} \\
\midrule
\multicolumn{8}{l}{\textit{Open-or-Close}} \\
Close Double Door
& 88.5
& 80.0
& \meanstd{88.0}{7.2}
& \meanstd{60.0}{5.3}
& \meanstd{90.0}{2.0}
& \meanstd{57.3}{25.7}
& \meanstd{72.0}{6.9} \\
Close Drawer
& 100.0
& 100.0
& \meanstd{92.7}{8.1}
& \meanstd{68.0}{8.0}
& \meanstd{100.0}{0.0}
& \meanstd{77.3}{39.3}
& \meanstd{98.7}{2.3} \\
Close Single Door
& 96.0
& 95.0
& \meanstd{96.0}{4.0}
& \meanstd{89.3}{9.5}
& \meanstd{97.3}{3.1}
& \meanstd{73.3}{11.5}
& \meanstd{68.7}{1.2} \\
Open Double Door
& 39.0
& 25.0
& \meanstd{52.0}{5.3}
& \meanstd{94.0}{6.0}
& \meanstd{75.3}{3.1}
& \meanstd{32.0}{5.3}
& \meanstd{10.7}{2.3} \\
Open Drawer
& 81.1
& 95.0
& \meanstd{58.0}{8.7}
& \meanstd{88.0}{3.5}
& \meanstd{55.3}{5.8}
& \meanstd{42.7}{1.2}
& \meanstd{36.0}{14.4} \\
Open Single Door
& 81.5
& 90.0
& \meanstd{68.7}{3.1}
& \meanstd{82.7}{8.1}
& \meanstd{78.7}{7.0}
& \meanstd{48.0}{18.3}
& \meanstd{50.7}{6.1} \\
\midrule
\multicolumn{8}{l}{\textit{Others}} \\
Coffee Press Button
& 98.5
& 100.0
& \meanstd{98.0}{2.0}
& \meanstd{32.0}{2.0}
& \meanstd{100.0}{0.0}
& \meanstd{20.0}{8.0}
& \meanstd{74.7}{6.1} \\
Coffee Serve Mug
& 63.5
& 85.0
& \meanstd{74.7}{9.0}
& \meanstd{73.3}{4.2}
& \meanstd{81.3}{7.0}
& \meanstd{25.3}{9.2}
& \meanstd{23.3}{9.9} \\
Coffee Setup Mug
& 31.0
& 30.0
& \meanstd{32.0}{10.6}
& \meanstd{79.3}{13.3}
& \meanstd{30.7}{1.2}
& \meanstd{14.7}{12.2}
& \meanstd{4.0}{0.0} \\
Turn Off Microwave
& 96.0
& 95.0
& \meanstd{99.3}{1.2}
& \meanstd{92.7}{4.2}
& \meanstd{99.3}{1.2}
& \meanstd{72.0}{16.0}
& \meanstd{90.7}{4.6} \\
Turn Off Sink Faucet
& 93.5
& 100.0
& \meanstd{93.3}{6.4}
& \meanstd{85.3}{7.6}
& \meanstd{98.0}{2.0}
& \meanstd{74.7}{19.7}
& \meanstd{89.3}{6.1} \\
Turn Off Stove
& 31.0
& 25.0
& \meanstd{27.3}{8.1}
& \meanstd{93.3}{5.8}
& \meanstd{28.7}{2.3}
& \meanstd{22.7}{22.0}
& \meanstd{16.7}{4.2} \\
Turn On Microwave
& 91.5
& 95.0
& \meanstd{89.3}{8.1}
& \meanstd{92.7}{4.6}
& \meanstd{93.3}{4.2}
& \meanstd{53.3}{2.3}
& \meanstd{72.7}{6.4} \\
Turn On Sink Faucet
& 89.0
& 95.0
& \meanstd{84.0}{8.0}
& \meanstd{18.7}{9.0}
& \meanstd{96.0}{4.0}
& \meanstd{33.3}{18.5}
& \meanstd{61.3}{12.9} \\
Turn On Stove
& 76.5
& 85.0
& \meanstd{72.0}{4.0}
& \meanstd{60.0}{3.5}
& \meanstd{70.7}{2.3}
& \meanstd{33.3}{11.5}
& \meanstd{41.3}{10.1} \\
Turn Sink Spout
& 87.0
& 80.0
& \meanstd{97.3}{3.1}
& \meanstd{93.3}{4.2}
& \meanstd{93.3}{5.8}
& \meanstd{45.3}{2.3}
& \meanstd{61.3}{6.1} \\
\midrule
\textbf{Pick-and-Place}
& 43.2
& 53.0
& \meanstd{39.4}{1.0}
& \meanstd{47.0}{4.4}
& \bestmeanstd{49.8}{2.3}
& \meanstd{14.0}{1.7}
& \bestmeanstd{15.3}{1.4} \\
\textbf{Open-or-Close}
& 81.0
& 80.8
& \meanstd{75.9}{3.4}
& \meanstd{80.3}{2.3}
& \bestmeanstd{82.8}{1.3}
& \meanstd{55.1}{7.4}
& \bestmeanstd{56.1}{3.6} \\
\textbf{Others}
& 75.8
& 79.0
& \meanstd{76.7}{0.9}
& \meanstd{72.1}{2.0}
& \bestmeanstd{79.1}{1.5}
& \meanstd{39.5}{0.9}
& \bestmeanstd{53.5}{2.4} \\
\textbf{Average}
& 66.2
& 70.8
& \meanstd{64.1}{1.6}
& \meanstd{65.8}{0.8}
& \bestmeanstd{70.3}{1.7}
& \meanstd{34.9}{2.0}
& \bestmeanstd{41.4}{1.4} \\
\bottomrule
\end{tabular}%
}
\end{table*}

%Uncomment after qualitative assets are prepared.
\subsection{Qualitative Results}
\label{app:qualitative_results}

\begin{figure}[htbp]
    \centering
    \includegraphics[width=0.95\textwidth]{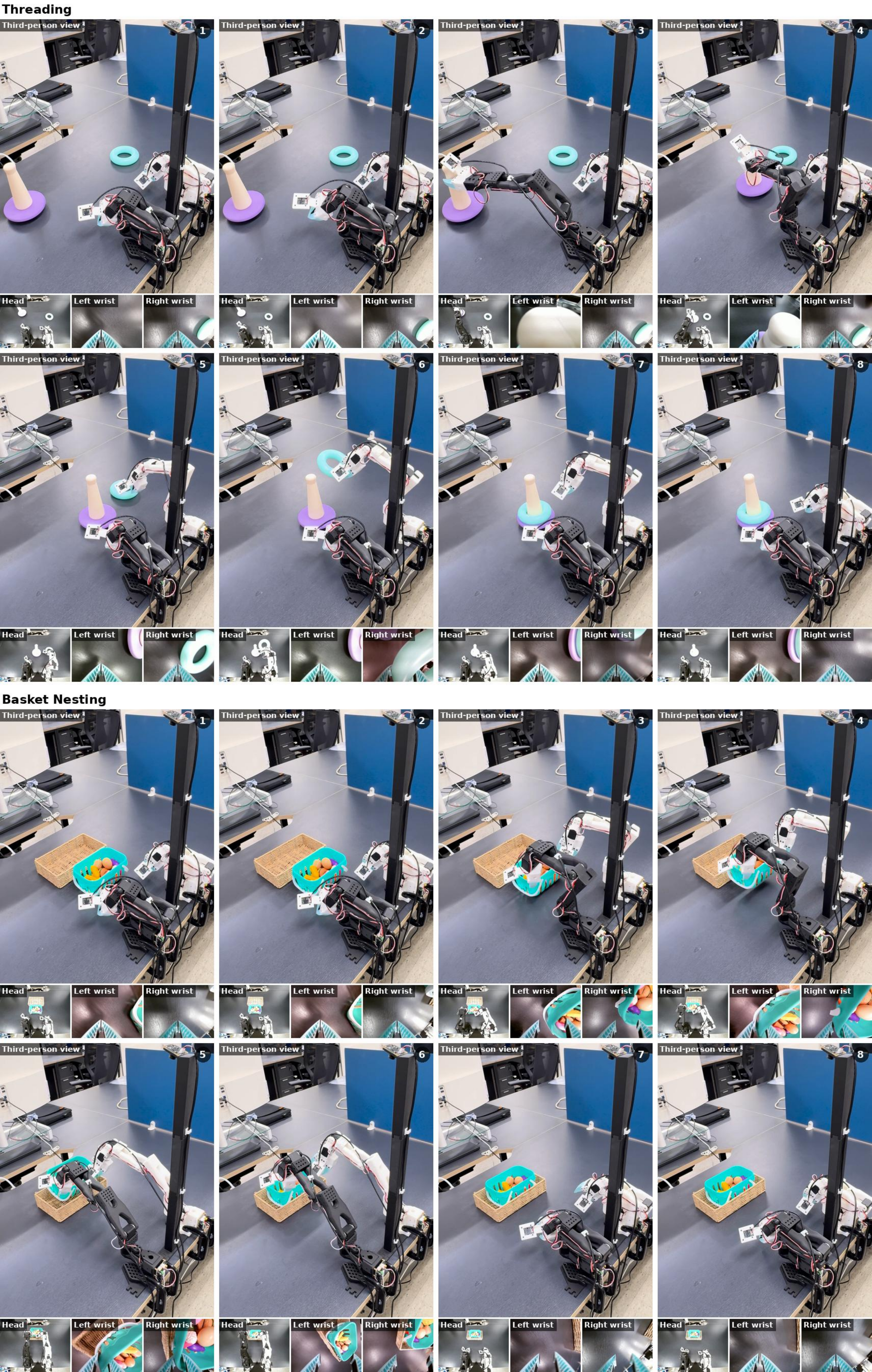}
    \caption{\textbf{Representative real-world rollouts on Threading and Basket Nesting.}
    The filmstrips show successful execution across the third-person, head, and wrist-camera views.}
    \label{fig:realworld-qualitative-short}
\end{figure}

\begin{figure}[htbp]
    \centering
    \includegraphics[width=0.95\textwidth]{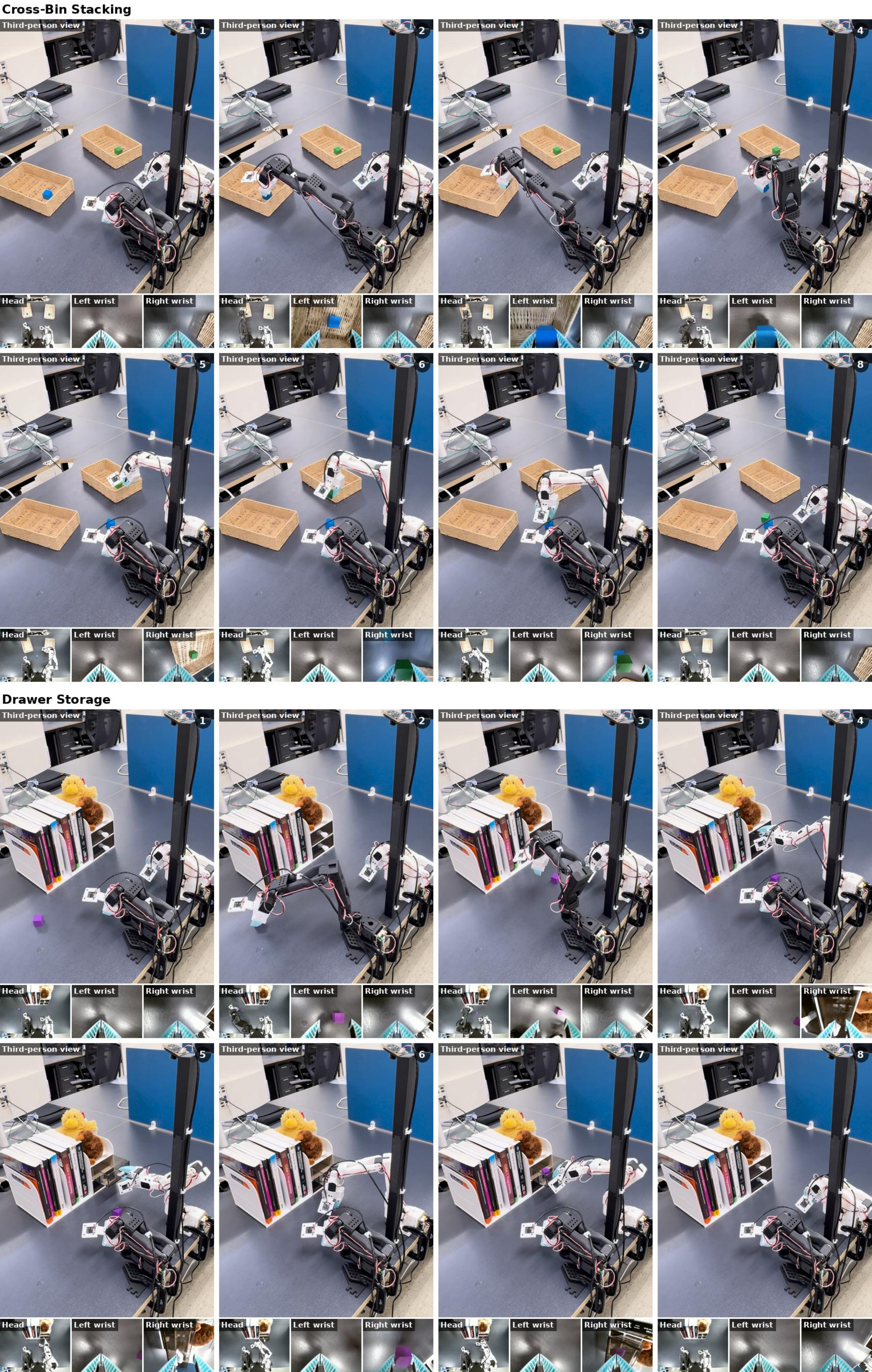}
    \caption{\textbf{Representative real-world rollouts on Cross-Bin Stacking and Drawer Storage.}
    The examples illustrate coordinated execution across the multiple stages required by the two longer-horizon tasks.}
    \label{fig:realworld-qualitative-long}
\end{figure}

We provide representative real-world rollouts for all four tasks in
\Cref{fig:realworld-qualitative-short,fig:realworld-qualitative-long}.
Each filmstrip shows the third-person view together with the head and two wrist-camera views.
The complete rollout videos are provided on the project page.

%%%%%%%%%%%%%%%%%%%%%%%%%%%%%%%%%%%%%%%%%%%%%%%%%%%%%%%%%%%%%%%%%%%%%%%%%%%%%%%%
%%%%%%%%%%%%%%%%%%%%%%%%%%%%%%%%%%%%%%%%%%%%%%%%%%%%%%%%%%%%%%%%%%%%%%%%%%%%%%%%
\section{Limitations and Future Work}
\label{app:limitations}

\method{} inherits its notion of intent from a teacher VLM, so the semantics available to the policy are bounded by what that teacher understands about physical behavior.
The bound is consequential rather than cosmetic.
A teacher without embodied grounding yields targets with almost no sample-dependent direction, and the resulting policy falls below the action-only baseline (\Cref{tab:teacher-policy-transfer,tab:teacher-target-geometry}).
This property can be measured before training, but the framework still has no notion of an intent being wrong and no mechanism that revises a target when execution contradicts it.
Removing this ceiling requires intent that is validated against execution outcomes or acquired through the policy's own interaction rather than read from a fixed interpreter.

A second limit is that intent is recovered rather than chosen.
The deployed policy reconstructs one deterministic intent at each query and immediately consumes it.
Our interventions show that this state is addressable (\Cref{app:validating_intent}), yet the policy never uses that handle.
It cannot hold a distribution over valid objectives when the stage is ambiguous, maintain an intent across a rollout, or accept a correction stated as what it should be trying to do.
Treating intent as an interface for a planner, a human, or execution feedback is the extension we consider most consequential.

Finally, intent is read from single windows of demonstrations that succeeded.
Manipulation objectives are nested, and failed or corrected episodes state the objective most explicitly while accounting for much of the experience a robot can collect.
Hierarchical intent and intent distilled from unsuccessful behavior are natural next steps.

%%%%%%%%%%%%%%%%%%%%%%%%%%%%%%%%%%%%%%%%%%%%%%%%%%%%%%%%%%%%%%%%%%%%%%%%%%%%%%%%
%%%%%%%%%%%%%%%%%%%%%%%%%%%%%%%%%%%%%%%%%%%%%%%%%%%%%%%%%%%%%%%%%%%%%%%%%%%%%%%%

\section{Use of Large Language Models}
\label{app:llm_use}

We used large-language-model-based tools, including Claude Code and Codex CLI, to assist with code navigation, experiment orchestration, and language editing.
All methodological decisions, implementation changes, experimental procedures, and reported results were reviewed and verified by the authors.
The Cosmos-Reason2 teacher used to construct training targets is a component of the proposed method and is distinct from the tools used during manuscript and code preparation.

\end{document}